\documentclass[lettersize,journal]{IEEEtran}
\usepackage{algorithmic}
\usepackage{color,array}
\usepackage[colorlinks,urlcolor=blue,linkcolor=blue,citecolor=blue]{hyperref}
\usepackage{textcomp}
\usepackage{stfloats}
\usepackage{url}
\usepackage{verbatim}
\usepackage{amsfonts,amsthm,amsmath,amssymb}
\usepackage{cleveref}
\usepackage{subfigure}
\usepackage{xurl}
\usepackage{xparse}
\usepackage{flushend}
\usepackage{wrapfig}
\usepackage{xfrac}
\usepackage{graphicx}
\usepackage{comment}
\usepackage{booktabs}
\usepackage{multirow}
\usepackage{pifont}
\usepackage{bbding}
\usepackage{float}
\usepackage[table,xcdraw]{xcolor}
\def\BibTeX{{\rm B\kern-.05em{\sc i\kern-.025em b}\kern-.08em
    T\kern-.1667em\lower.7ex\hbox{E}\kern-.125emX}}
\usepackage{balance}

\newcommand{\flag}{0}

\ifnum\flag=2
  \newcommand{\revise}[1]{\textcolor{blue}{#1}}
\else
    \newcommand{\revise}[1]{#1}
\fi

\ifnum\flag=2
  \newcommand{\modify}[1]{\textcolor{red}{#1}}
\else
    \newcommand{\modify}[1]{#1}
\fi

\ifnum\flag=3
  \newcommand{\rone}[1]{\textcolor{red}{#1}}
\else
    \newcommand{\rone}[1]{#1}
\fi

\ifnum\flag=4
  \newcommand{\rtwo}[1]{\textcolor{red}{#1}}
\else
    \newcommand{\rtwo}[1]{#1}
\fi

\ifnum\flag=5
  \newcommand{\rthree}[1]{\textcolor{red}{#1}}
\else
    \newcommand{\rthree}[1]{#1}
\fi

\newcommand{\redbf}[1]{\textbf{\textcolor{red}{#1}}}

\newcommand{\etal}{\textit{~et~al}}

\begin{document}
\title{Balancing Efficiency and Restoration: Lightweight Mamba-Based Model for CT Metal Artifact Reduction}
\author{Weikai Qu, Sijun Liang, Xianfeng Li, Cheng Pan, An Yan, \\Ahmed Elazab, Shanzhou Niu, Dong Zeng*, Xiang Wan and Changmiao Wang*
\thanks{Manuscript created July, 2025; This work was supported in part by Guangxi Science and Technology Program (No. FN2504240022), Guangxi Key R\&D Project (No. AB24010167), Guangdong Basic and Applied Basic Research Foundation (No. 2025A1515011617). This work was also supported in part by the National Key R\&D Program of China under Grants 2024YFA1012000 and 2024YFC2417804, the National Natural Science Foundation of China under Grant 62571228, Beijing Natural Science Foundation under Grant Z250002, and Guangdong Basic and Applied Basic Research Foundation under Grant 2024A1515140146. \textit{(Corresponding author: Dong Zeng and Changmiao Wang)}}
\thanks{This work did not involve human subjects or animals in its research.}
\thanks{Weikai Qu is with New York University, America (e-mail: weikaiqu@nyu.edu).}
\thanks{Sijun Liang is with BaiyunPort, China (e-mail: chrisliang97@gmail.com).}
\thanks{Xianfeng Li is with University of Technology Sydney, Australia (e‑mail: xianfeng.li2003@gmail.com).}
\thanks{Cheng Pan is with Sanda University, China (e-mail: panc@sandau.edu.cn).}
\thanks{An Yan is with Shan Dong Xiehe College, China (e-mail: Yanan@sdxiehe.edu.cn).}
\thanks{Ahmed Elazab is with Tsinghua University, China (e-mail: ahmed.elazab@yahoo.com).}
\thanks{Shanzhou Niu is with Gannan Normal University, China (e-mail: szniu@gnnu.edu.cn).}
\thanks{Dong Zeng is with Southern Medical University, China (e-mail: zd1989@smu.edu.cn).}
\thanks{Xiang Wan is with Shenzhen Research Institute of Big Data, China (e-mail: wanxiang@sribd.cn).}
\thanks{Changmiao Wang is with Shan Dong Xiehe College, China, and with Shenzhen Research Institute of Big Data, China (e-mail: cmwangalbert@gmail.com).}
}

\markboth{IEEE TRANSACTIONS ON RADIATION AND PLASMA MEDICAL SCIENCES, VOL. 0, NO. 0, JULY 2025}
{QU \MakeLowercase{\textit{et al.}}: Balancing Efficiency and Restoration: Lightweight Mamba-Based Model for CT Metal Artifact Reduction}

\maketitle

\begin{abstract}
In computed tomography imaging, metal implants frequently generate severe artifacts that compromise image quality and hinder diagnostic accuracy. \modify{There are three main challenges in the existing methods: the deterioration of organ and tissue structures, dependence on sinogram data, and an imbalance between resource use and restoration efficiency. Addressing these issues, we introduce MARMamba, which effectively eliminates artifacts caused by metals of different sizes while maintaining the integrity of the original anatomical structures of the image. Furthermore, this model only focuses on CT images affected by metal artifacts, thus negating the requirement for additional input data.} The model is a streamlined UNet architecture, which incorporates multi-scale Mamba (MS-Mamba) as its core module. Within MS-Mamba, a flip mamba block captures comprehensive contextual information by analyzing images from multiple orientations. Subsequently, the average maximum feed-forward network integrates critical features with average features to suppress the artifacts. This combination allows MARMamba to reduce artifacts efficiently. \modify{The experimental results demonstrate that our model excels in reducing metal artifacts, offering distinct advantages over other models. It also strikes an optimal balance between computational demands, memory usage, and the number of parameters, highlighting its practical utility in the real world.} The code of the presented model is available at: \url{https://github.com/RICKand-MORTY/MARMamba}.
\end{abstract}


\section{Introduction}
\IEEEPARstart{C}{omputed} tomography (CT) is widely used in clinical diagnostics to provide high-resolution images to detect and evaluate pathological conditions~\cite{indudonet}. However, metal implants (e.g., orthopedic devices, dental restorations) significantly degrade CT imaging quality due to artifacts caused by photon starvation, beam hardening, and scattering effects~\cite{review,review2}. As the prevalence of metallic implants increases, effectively reducing these metal artifacts and restoring the obscured details in CT images remains a critical but challenging task in medical imaging. 

Traditional metal artifact reduction (MAR) methods focus primarily on employing estimation strategies for corrupted sinogram regions~\cite{li,nmar}. However, these methods may introduce secondary blurring or artifacts. 
The rapid development of deep learning has significantly advanced MAR research, yielding numerous models that operate through either dual-domain processing (combining image-domain and sinogram-domain approaches) or specialized single-domain enhancement techniques.
Wang\etal. developed interpretable dual-domain networks~\cite{indudonet,indudonet_plus} alongside a rotation-equivariant convolutional neural network (CNN)~\cite{mepnet}, which leverages the rotational priors inherent to CT scanning to enhance dual-domain processing. \rone{In recent years, more sophisticated dual-domain frameworks have been developed to address the challenges of artifact suppression and image restoration in medical imaging. For instance, Wang et al. introduced IDOL-Net, a model that leverages a parallel and interactive dual-domain structure. This design facilitates effective information exchange between the two domains, maximizing the utilization of latent features for improved performance~\cite{IDOLNET}. Similarly, IRDNet represents another notable advancement, as it explicitly incorporates metal artifact features and applies a residual learning strategy. This iterative refinement process enhances both artifact suppression and image restoration quality~\cite{IRDNET}.}
\rone{However, despite these advancements, obtaining accurate and reliable data in clinical settings remains a significant hurdle. While threshold-based methods can be used to generate metal masks, their effectiveness is constrained by the difficulties of achieving precision in real-world clinical environments. This limitation continues to impede their broader adoption in practice~\cite{CBCT}}. Moreover, cross-domain processing significantly increases the computational burden, making it highly unfavorable for deployment.

In the realm of image domain only enhancement, Wang\etal. developed convolutional dictionary networks that take advantage of the sample-invariant features of CT images~\cite{acdnet} and further capitalized on the rotationally symmetrical structure of metal streak artifacts~\cite{oscnet,oscnet_plus}. \modify{LWCDNet utilizes weighted convolutional dictionary units and a multi-stage iterative network design to improve both the interpretability of the approach and its ability to manage metal artifacts of different sizes~\cite{LWCDNet}.}

\modify{In recent years, diffusion model~\cite{diffusion} techniques have emerged as a promising approach to image enhancement, providing innovative solutions in this field. Karageorgos\etal. applied an unconditional denoising diffusion probabilistic model (DDPM)~\cite{DDPM} to restore corrupted projection data~\cite{ddpm_mar}.
Liu\etal. combined a diffusion prior with dynamic weight masks to effectively balance prior information and likelihood~\cite{dudop}. Xia\etal. introduced a dual-domain DDPM method that simultaneously operates in both the sinogram and image domains, effectively eliminating artifacts while restoring image details~\cite{DD-DDPM}. They also employed the denoising diffusion implicit model (DDIM)~\cite{DDIM} technique to speed up the process. However, diffusion model-based methods require significant resources and inference time, limiting their practical application.}

Attention-based and Transformer-based~\cite{attention} models have also gained attention.
\modify{Zhu\etal. implemented a transformer efficiently integrated within a CycleGAN architecture~\cite{non-supervision-transformer}. In contrast, Xie\etal. used a dense Transformer to capture low-level features enriched with hierarchical information~\cite{dtecnet}. Deep learning methods often suffer from excessive smoothing during image restoration. To address this issue, MUPO-Net incorporates attention-based mechanisms and assigns weights to various regions based on metal artifact impacts~\cite{MUPO-Net}. This approach not only directs the restoration process in the sinogram domain but also utilizes the Transformer to improve feature representation. This dual strategy effectively mitigates the over-smoothing of reconstructed images, thereby maintaining the anatomical details of organs and tissues. Despite these advancements, the Transformer still faces challenges in effectively integrating local and global features.}

\modify{Recently, Mamba has gained attention as a promising sequence modeling architecture based on first-order linear time-invariant state space equations~\cite{mamba,mamba2}. It features a selective scanning mechanism where state transitions are dynamically influenced by the input, reimagining sequence modeling as a structured state evolution process. This design allows the model to capture long-range dependencies effectively without relying on explicit attention mechanisms. Mamba's computational efficiency is a significant advantage; it combines state updates and input mixing into a single operation, achieving linear time and space complexity. This efficiency ensures excellent scalability for processing long sequences and high-resolution data~\cite{mamba, s4}. Building on these strengths, recent research has adapted Mamba to visual domains, where it excels at jointly modeling local details and global contextual information while significantly reducing computational demands~\cite{visionmamba}. These attributes make Mamba particularly well-suited for CT metal artifact reduction, a task that requires modeling complex, long-range structural distortions caused by metallic implants. Mamba facilitates high-fidelity image reconstruction with fewer parameters and lower memory consumption, providing a practical and efficient solution for medical image restoration, especially in resource-limited clinical environments.}

\modify{In summary, existing mainstream methods encounter three primary challenges: they often compromise the anatomical integrity of organs or tissues, rely heavily on supplementary data like sinogram domains and masks, and suffer from uneven distribution of computational resources. To overcome these challenges and strike an optimal balance between restoration effectiveness and resource efficiency, we introduce MARMamba, a metal artifact reduction model specifically designed using the Mamba architecture.} MARMamba adopts a UNet-like~\cite{unet} architecture and utilizes the multi-scale Mamba (MS-Mamba) module as its core. Within MS-Mamba, the flip Mamba block (FMB) captures diverse orientation information through a multi-branch structure, while the average maximum feed-forward network (AMFN) combines max pooling and average pooling operations to effectively integrate features. By integrating the aforementioned techniques, our model successfully addresses the dependency on sinogram data found in existing research and alleviates the performance burdens associated with cross-domain fusion. \modify{Furthermore, MARMamba excels in balancing high restoration quality with efficient resource use, minimizing resource consumption while maintaining state-of-the-art performance in metal artifact reduction. These attributes highlight its practical applicability.} The key contributions of this work are as follows: 
\begin{enumerate}
    \item \modify{We present the FMB and AMFN modules, which are designed to integrate information from various feature map perspectives. These modules effectively combine detailed local features with global contextual information, improving the correction of metal artifact regions while preserving anatomical structures.}
    \item \modify{Our model focuses solely on processing images with metal artifacts, eliminating the need for sinogram-domain data and mask images. This approach aligns more closely with real-world conditions.}
    \item \modify{Our model achieves a balance between performance and effectiveness by maintaining high restoration quality with low resource consumption.}
\end{enumerate}

\section{Proposed Method}

\begin{figure*}[htbp]
\centering
\includegraphics[scale=0.25]{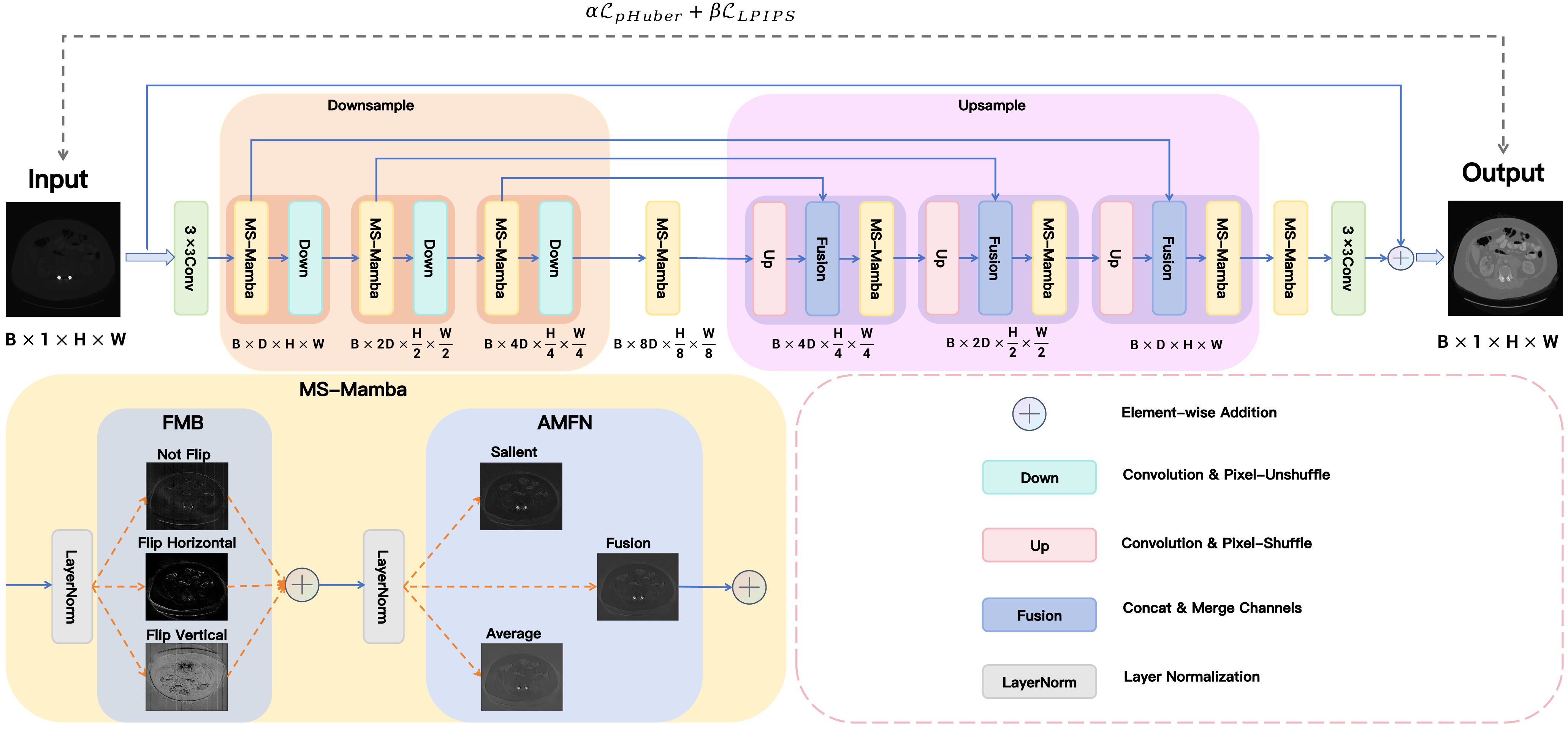}
\caption{\rthree{MARMamba backbone architecture. It consists of upsampling and downsampling components, each comprising three stages, with MS-Mamba serving as the core module for every stage.} 
}
\label{fig:backbone}
\end{figure*}

The MARMamba backbone network adopts a UNet-like architecture, as depicted in Fig.~\ref{fig:backbone}. The model's input dimension is specified as \( B \times 1 \times H \times W \), where \( B \) represents the batch size, and \( H \) and \( W \) denote the image's height and width, respectively. Instead of using patch embedding, we apply a \(3 \times 3\) convolutional layer to increase the input feature maps' channel count from 1 to \( D \), with \( D \) being a configurable parameter. The feature maps then undergo a series of downsampling stages, doubling the number of channels while halving the height and width at each step. This is followed by upsampling stages, where the channel count is halved, and the height and width are doubled. Finally, a \(3 \times 3\) convolutional layer reduces the channel count from \( D \) back to 1. This layer is followed by an element-wise addition with the original input to guide the model in learning artifact features. As a result, the final output image, free of artifacts, retains the same dimensions as the input. Within the MARMamba backbone network, the core module is MS-Mamba. This module uses the FMB to integrate contexts and the AMFN to aggregate features. By combining the context-sampling mechanism with the MS-Mamba module, the model effectively learns artifact-related features at multiple scales and leverages Mamba's context-binding capability.

\subsection{Multi-Scale Mamba}

\begin{figure*}[h]
\centering
\includegraphics[scale=0.4]{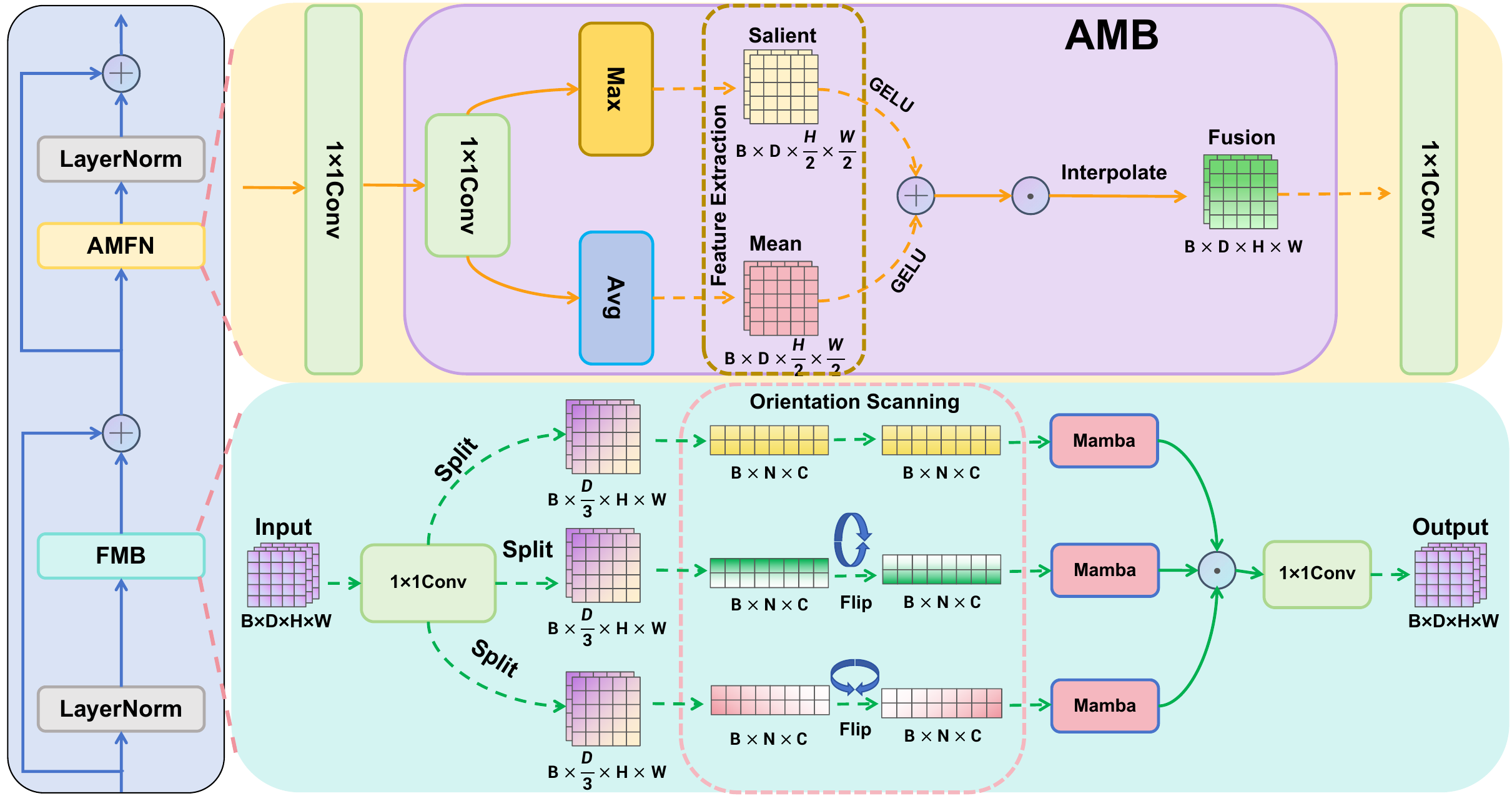}
\caption{\rthree{Internal structure of MS-Mamba, which consists of two components: FMB and AMFN. The right side displays the internal details of those two modules.}} 
\label{fig:block}
\end{figure*}

Fig.~\ref{fig:block} illustrates the internal structure of MS-Mamba, which consists of two parts, FMB and AMFN, each with a normalization layer and skip connections. Although Mamba excels in handling sequential data with long contextual dependencies, it is generally less effective for image processing tasks due to the lack of a specific order in image pixels~\cite{visionmamba,mambaout}. To address this limitation, we developed the FMB, which utilizes multiple Mamba instances, as illustrated in the lower right region of Fig.~\ref{fig:block}. In the FMB, Mamba processes the same feature map from multiple directions to capture features from various perspectives.

Initially, the input with dimensions \( B \times 1 \times H \times W \) is subjected to a \(1 \times 1\) convolution and then split into three parts along the channel dimension. Each part transforms a feature map of dimension \(B \times N \times C\), where \(N\) corresponds to \(H \times W\). The feature map's first part is directly processed by Mamba. In contrast, the second and third parts are first flipped vertically and horizontally, respectively, before being fed through Mamba. \rone{The two flipped feature maps provide additional information to enhance the unflipped feature map, effectively emphasizing artifact regions. These flipped maps are utilized as weights and combined with the unflipped map through element-wise multiplication. To complete the process, a \(1 \times 1\) convolution is applied to adjust and restore the channel count.} This approach enables FMB to extract orientation-specific information, thus enhancing Mamba's effectiveness in image-processing tasks. \rone{The AMFN module within the MS-Mamba framework strengthens feature learning by integrating max and average pooling operations. Additionally, the Average Maximum Block (AMB) captures both dominant and comprehensive image features, further enhancing feature representation through weight fusion achieved via element-wise multiplication.}

\subsection{Loss Function}

We utilize the Pseudo-Huber metric family~\cite{cm_improved} as the absolute loss function for our model. This metric is defined as follows: 

\begin{equation}
  \mathcal{L}_{pHuber}(\textbf{Y},\hat{\textbf{I}}) = \sqrt{\| \textbf{Y},\hat{\textbf{I}} \|^2_2 + c^2} - c.
\end{equation}
where \(\hat{\textbf{I}}\) denotes the output images from MARMamba, \(\textbf{Y}\) represent the ground truth images, and \(c\) be a weighting coefficient.
While the absolute loss function is effective, it may not fully capture the nuances that are perceptually significant to human vision. To address this, we also incorporate the LPIPS~\cite{lpips} metric as the perceptual loss function. LPIPS leverages a pre-trained deep learning model to assess perceptual image quality, aligning more closely with human visual perception. These two loss functions are combined with adjustable weights to balance their contributions, using the coefficients \(\alpha\) and \(\beta\). This combination allows us to fine-tune the model's sensitivity to both objective and perceptual quality measures, as given below:

\begin{equation}
  \mathcal{L}(\textbf{Y},\hat{\textbf{I}}) = \alpha\mathcal{L}_{pHuber}(\textbf{Y},\hat{\textbf{I}})+\beta\mathcal{L}_{LPIPS}(\textbf{Y},\hat{\textbf{I}}).
\end{equation}

\section{Experiments and Results}

\subsection{Dataset and Implementation}

\textbf{Dataset and Preprocessing.} We train our model using the training set of a synthetic dataset and evaluate its performance on both the test set of the synthetic dataset and a real-world dataset, detailed below. For real scene inference, we apply thresholding to segment the mask, excise the metal region from the image, input the modified image into the model, and subsequently reintroduce the metal region into the output.

\textbf{The SynDeepLesion Dataset~\cite{indudonet}.}  SynDeepLesion is a synthetic dataset derived from 1,200 images selected from the DeepLesion dataset~\cite{deeplesion}. It is divided into two subsets: a training set containing 1,000 image pairs and a test set comprising 200 image pairs. In the test set, metal sizes range as follows: [2061, 890, 881, 451, 254, 124, 118, 112, 53, 35]. For analysis, we categorize these sizes into four groups: \textbf{large} ([2061, 890, 881]), \textbf{medium} ([451, 254]), \textbf{small} ([124, 118, 112]), and \textbf{tiny} ([53, 35]).

\textbf{The CLINIC-metal Dataset~\cite{CTPelvic1K}.} The CLINIC-metal dataset consists of postoperative medical images containing metal artifacts, collected from an orthopedic hospital. This dataset serves to evaluate the model's ability to restore images with metal artifacts in real-world clinical scenarios.

\textbf{Implementation Details.}
The channels count \( D \) is set to 12. The number of MS-Mamba blocks in each layer is configured sequentially as [1, 2, 2, 4, 2, 2, 1, 1]. In the loss function, we set $\alpha$ to 0.8, $\beta$ to 0.2 and $c$ to 0.03. The model was implemented using PyTorch 2.3.0 with CUDA 12.1 and trained on a single NVIDIA RTX 3090 GPU. 

\rone{For training, we implemented the Adam optimizer~\cite{adam} alongside a CosineAnnealingLR learning rate schedule~\cite{CosineAnnealingLR}. Larger image sizes improve the model's ability to capture detailed features, but GPU memory limitations often require smaller batch sizes, leading to potential instability during training. To mitigate this, we employed a progressive training strategy~\cite{restormer}. This approach begins with large batches of smaller images and gradually shifts to larger images with reduced batch sizes, stabilizing the training while enhancing model performance. The training process is divided into three phases: 100,000 iterations with \(256 \times 256\) images and a batch size of 8, followed by 200,000 iterations with \(336 \times 336\) images and a batch size of 4, and finally, 20,000 iterations with \(416 \times 416\) images and a batch size of 2.} \rone{Across all three training phases, the initial learning rate was set to 0.0002. The maximum number of iterations (\(T_{\text{max}}\)) was configured at 1,000, while the minimum learning rate was fixed at \(1 \times 10^{-8}\). The total training process was completed in 27.31 hours.} 

\rone{We developed MARformer using the MARformer-B configuration~\cite{marformer}. To ensure the parameters remain consistent, we adjusted the output dimension of the initial convolutional layer to match that of MARMamba. The training setup follows the MARformer paper, with the exception of the iteration count, which was adjusted to align with MARformer specifications.} In addition, we replaced MS-Mamba with CNN combined with ViT~\cite{vit} and PVT v2~\cite{pvtv2} to compare the effects of Mamba-based and Transformer-based models. These models, named MARViT and MARPVT respectively, use the same parameter settings as MARMamba, with attention heads set to [4, 6, 6, 8, 6, 6, 6, 4], and MARPVT's spatial reduction ratio set to 7. The three models above are trained using the same strategy as MARMamba.

\subsection{Comparative Experimental Analysis}

\begin{table*}[]
\caption{\rone{Quantitative comparison experiments. Evaluate the effectiveness of metal artifact restoration for \textbf{large} and \textbf{medium} sizes using the PSNR($\uparrow$), SSIM($\uparrow$), RMSE($\downarrow$), LPIPS($\downarrow$) metrics \textbf{(Non-metallic area)}. Each grid cell represents the mean $\pm$ standard deviation. The optimal mean values highlighted in \textbf{\color{red}{bold red}}.}}
\centering
\resizebox{\textwidth}{!}{
\begin{tabular}{ccccc|cccc}
\hline
\multirow{2}{*}{Model} & \multicolumn{4}{c|}{Large} & \multicolumn{4}{c}{Medium} \\
 & PSNR & SSIM & RMSE & LPIPS & PSNR & SSIM & RMSE & LPIPS \\ \hline
InDuDoNet~\cite{indudonet} & 39.77 ± 3.60 & 0.9881 ± 0.0051 & 3.30 ± 1.26 & 0.0576 ± 0.0176 & 41.72 ± 4.19 & 0.9926 ± 0.0022 & 2.76 ± 1.58 & 0.0430 ± 0.0111 \\
InDuDoNet+~\cite{indudonet_plus} & 38.29 ± 3.36 & 0.9858 ± 0.0049 & 3.93 ± 1.97 & 0.0639 ± 0.0164 & 40.72 ± 3.74 & 0.9904 ± 0.0029 & 3.02 ± 1.54 & 0.0501 ± 0.0122 \\
OSCNet~\cite{oscnet} & 41.14 ± 2.83 & 0.9910 ± 0.0023 & 2.77 ± 1.18 & 0.0476 ± 0.0128 & 41.08 ± 4.24 & 0.9923 ± 0.0023 & 3.00 ± 1.88 & 0.0401 ± 0.0112 \\
OSCNet+~\cite{oscnet_plus} & 41.51 ± 2.96 & 0.9911 ± 0.0023 & 2.66 ± 1.17 & 0.0474 ± 0.0126 & 41.88 ± 3.88 & 0.9928 ± 0.0019 & 2.69 ± 1.61 & 0.0400 ± 0.0112 \\
MEPNet~\cite{mepnet} & 38.28 ± 3.33 & 0.9852 ± 0.0073 & 3.90 ± 1.55 & 0.0648 ± 0.0194 & 42.90 ± 3.78 & 0.9922 ± 0.0021 & 2.36 ± 1.27 & 0.0488 ± 0.0144 \\
DICDNet~\cite{dicdnet} & 39.91 ± 3.78 & 0.9890 ± 0.0033 & 3.37 ± 2.13 & 0.0577 ± 0.0146 & 40.93 ± 3.72 & 0.9917 ± 0.0020 & 2.96 ± 1.65 & 0.0473 ± 0.0121 \\
ACDNet~\cite{acdnet} & 39.90 ± 2.72 & 0.9903 ± 0.0027 & 3.17 ± 1.18 & 0.0500 ± 0.0131 & 40.89 ± 3.84 & 0.9926 ± 0.0017 & 2.99 ± 1.63 & 0.0417 ± 0.0110 \\
DuDoDp~\cite{dudop} & 43.47 ± 1.73 & 0.9886 ± 0.0032 & 2.03 ± 0.42 & 0.0622 ± 0.0092 & 44.45 ± 1.30 & 0.9905 ± 0.0022 & 1.80 ± 0.27 & 0.0552 ± 0.0082 \\
MARformer~\cite{marformer} & 44.88 ± 4.12 & 0.9912 ± 0.0051 & 1.90 ± 0.93 & 0.0643 ± 0.0245 & 47.74 ± 3.46 & 0.9947 ± 0.0018 & 1.31 ± 0.50 & 0.0464 ± 0.0128 \\
MARViT (ours) & 46.26 ± 1.47 & 0.9929 ± 0.0017 & 1.46 ± 0.26 & 0.0674 ± 0.0116 & 46.57 ± 1.95 & 0.9936 ± 0.0012 & 1.43 ± 0.32 & 0.0681 ± 0.0092 \\
MARPVT (ours) & 46.62 ± 1.09 & 0.9938 ± 0.0016 & 1.40 ± 0.18 & 0.0411 ± 0.0103 & 47.15 ± 2.45 & 0.9946 ± 0.0014 & 1.36 ± 0.37 & 0.0356 ± 0.0086 \\
MARMamba (ours) & \redbf{47.60} ± 1.65 & \redbf{0.9943} ± 0.0014 & \redbf{1.26} ± 0.23 & \redbf{0.0368} ± 0.0092 & \redbf{47.90} ± 3.20 & \redbf{0.9948} ± 0.0014 & \redbf{1.28} ± 0.45 & \redbf{0.0306} ± 0.0077 \\ \hline
\end{tabular}
}
\label{table:no_metal_large_medium}
\end{table*}

\begin{table*}[]
\caption{\rone{Quantitative comparison experiments. Evaluate the effectiveness of metal artifact restoration for \textbf{small} and \textbf{tiny} sizes using the PSNR($\uparrow$), SSIM($\uparrow$), RMSE($\downarrow$), LPIPS($\downarrow$) metrics \textbf{(Non-metallic area)}. Each grid cell represents the mean $\pm$ standard deviation. The optimal mean values highlighted in \textbf{\color{red}{bold red}}.}}
\centering
\resizebox{\textwidth}{!}{
\begin{tabular}{ccccc|cccc}
\hline
\multirow{2}{*}{Model} & \multicolumn{4}{c|}{Small} & \multicolumn{4}{c}{Tiny} \\
 & PSNR & SSIM & RMSE & LPIPS & PSNR & SSIM & RMSE & LPIPS \\ \hline
InDuDoNet~\cite{indudonet} & 46.37 ± 2.17 & 0.9949 ± 0.0011 & 1.47 ± 0.41 & 0.0375 ± 0.0092 & 46.72 ± 2.13 & 0.9953 ± 0.0008 & 1.41 ± 0.40 & 0.0364 ± 0.0087 \\
InDuDoNet+~\cite{indudonet_plus} & 46.72 ± 2.10 & 0.9940 ± 0.0013 & 1.41 ± 0.38 & 0.0414 ± 0.0101 & 47.42 ± 1.80 & 0.9946 ± 0.0009 & 1.29 ± 0.29 & 0.0392 ± 0.0096 \\
OSCNet~\cite{oscnet} & 46.24 ± 2.46 & 0.9948 ± 0.0012 & 1.51 ± 0.46 & 0.0339 ± 0.0090 & 47.17 ± 2.26 & 0.9952 ± 0.0009 & 1.35 ± 0.44 & 0.0325 ± 0.0091 \\
OSCNet+~\cite{oscnet_plus} & 46.49 ± 2.45 & 0.9948 ± 0.0012 & 1.47 ± 0.44 & 0.0339 ± 0.0093 & 47.37 ± 2.33 & 0.9952 ± 0.0009 & 1.32 ± 0.43 & 0.0325 ± 0.0092 \\
MEPNet~\cite{mepnet} & 47.79 ± 1.79 & 0.9943 ± 0.0012 & 1.24 ± 0.28 & 0.0430 ± 0.0122 & 48.16 ± 1.70 & 0.9946 ± 0.0011 & 1.19 ± 0.28 & 0.0413 ± 0.0120 \\
DICDNet~\cite{dicdnet} & 45.88 ± 2.38 & 0.9942 ± 0.0012 & 1.57 ± 0.46 & 0.0400 ± 0.0102 & 46.52 ± 2.19 & 0.9945 ± 0.0009 & 1.45 ± 0.43 & 0.0388 ± 0.0098 \\
ACDNet~\cite{acdnet} & 43.64 ± 3.07 & 0.9947 ± 0.0011 & 2.08 ± 0.75 & 0.0353 ± 0.0095 & 43.96 ± 3.35 & 0.9950 ± 0.0009 & 2.03 ± 0.79 & 0.0335 ± 0.0092 \\
DuDoDp~\cite{dudop} & 45.84 ± 1.08 & 0.9928 ± 0.0012 & 1.53 ± 0.19 & 0.0498 ± 0.0068 & 45.98 ± 1.03 & 0.9930 ± 0.0010 & 1.50 ± 0.18 & 0.0496 ± 0.0067 \\
MARformer~\cite{marformer} & 50.01 ± 1.40 & 0.9959 ± 0.0008 & 0.95 ± 0.18 & 0.0445 ± 0.0112 & 51.24 ± 1.05 & 0.9963 ± 0.0005 & 0.82 ± 0.14 & 0.0425 ± 0.0114 \\
MARViT (ours) & 48.80 ± 0.84 & 0.9948 ± 0.0007 & 1.08 ± 0.11 & 0.0686 ± 0.0103 & 49.21 ± 0.70 & 0.9950 ± 0.0004 & 1.03 ± 0.09 & 0.0666 ± 0.0085 \\
MARPVT (ours) & 49.95 ± 0.91 & \redbf{0.9960} ± 0.0007 & 0.95 ± 0.10 & 0.0335 ± 0.0077 & 50.60 ± 0.68 & 0.9962 ± 0.0005 & 0.88 ± 0.07 & 0.0323 ± 0.0074 \\
MARMamba (ours) & \redbf{50.91} ± 1.33 & \redbf{0.9960} ± 0.0008 & 0.86 ± 0.13 & \redbf{0.0297} ± 0.0067 & \multicolumn{1}{l}{\redbf{52.03} ± 0.95} & \multicolumn{1}{l}{\redbf{0.9964} ± 0.0005} & \multicolumn{1}{l}{\redbf{0.75} ± 0.09} & \multicolumn{1}{l}{\redbf{0.0263} ± 0.0057} \\ \hline
\end{tabular}
}
\label{table:no_metal_small_tiny}
\end{table*}

\rone{In our comparative experiments, we quantitatively evaluated the SynDeepLesion dataset~\cite{indudonet} using metrics such as PSNR~\cite{psnr}, SSIM~\cite{ssim}, RMSE, and LPIPS~\cite{lpips}. Our assessment focused specifically on areas outside the metal regions for each group of metal artifact images within the test set. The results are presented in Tables~\ref{table:no_metal_large_medium},~\ref{table:no_metal_small_tiny}.}

\rone{In the evaluation of large- and medium-scale artifact removal (Table~\ref{table:no_metal_large_medium}), physics-based methods demonstrate limited effectiveness across all quantitative metrics. For large artifact regions, PSNR values mostly range between 38–41 dB, with RMSE values between 2.7 and 3.9. The high standard deviations observed further indicate a lack of stability and suboptimal restoration quality. Although performance improves moderately for medium-scale artifacts, it remains below the levels achieved by more advanced approaches. In contrast, the diffusion-based model DuDoDp~\cite{dudop} delivers better results compared to physics-based methods, offering improved stability. Transformer-based architectures consistently outperform their physics-driven counterparts, with MARformer exhibiting strong performance on medium-scale artifacts, achieving a PSNR of 47.74 dB and reducing RMSE to 1.31. Further analysis reveals that both MARViT and MARPVT provide high numerical accuracy and stability. Among these, MARPVT achieves a better balance between perceptual quality and numerical performance, aligning more closely with human visual assessment.}

\rone{Our proposed MARMamba model demonstrates significant superiority across both large and medium scales. In large-scale scenarios, MARMamba exceeds the performance of MARPVT by 0.98 dB in PSNR and reduces RMSE by 0.14. At the medium scale, it achieves a further gain of 0.75 dB in PSNR, highlighting its strong recovery capability for various artifact sizes. While MARMamba consistently delivers the best average performance across all metrics, it does exhibit slightly higher standard deviations compared to MARPVT and MARViT. This indicates that although the model excels in both accuracy and perceptual fidelity, there is room for improvement to enhance stability and reduce variability across different test cases.}

\rone{An analysis of artifact removal at small and tiny scales (Table~\ref{table:no_metal_small_tiny}) reveals trends consistent with those observed in the large- and medium-scale experiments. Physics-based methods continue to demonstrate limitations in both numerical accuracy and stability, while Transformer-based approaches consistently outperform both physics-driven and diffusion models. Among these, MARformer exhibits notable numerical accuracy, whereas MARPVT strikes a more effective balance between perceptual quality and stability. MARMamba, however, maintains its overall superiority across these smaller scales. Although its standard deviation is slightly higher than that of certain Transformer-based models, it remains low, highlighting the model’s robustness and reliability across different artifact sizes. These findings extend and reinforce the advantages previously demonstrated by MARMamba in large- and medium-scale evaluations, underscoring its comprehensive performance across all scales.}

\begin{table*}[]
\caption{\rone{Quantitative comparison experiments. Evaluate the effectiveness of metal artifact restoration for \textbf{large} and \textbf{medium} sizes using the PSNR($\uparrow$), SSIM($\uparrow$), RMSE($\downarrow$), LPIPS($\downarrow$) metrics \textbf{(metal-included region)}. Each grid cell represents the mean $\pm$ standard deviation. The optimal mean values highlighted in \redbf{bold red font}.}}
\centering
\resizebox{\textwidth}{!}{
\begin{tabular}{ccccc|cccc}
\hline
\multirow{2}{*}{Model} & \multicolumn{4}{c|}{Large} & \multicolumn{4}{c}{Medium} \\
 & PSNR & SSIM & RMSE & LPIPS & PSNR & SSIM & RMSE & LPIPS \\ \hline
InDuDoNet~\cite{indudonet} & 34.62 ± 2.95 & 0.9801 ± 0.0074 & 5.83 ± 1.93 & 0.0708 ± 0.0214 & 37.54 ± 3.47 & 0.9893 ± 0.0036 & 4.31 ± 2.21 & 0.0485 ± 0.0126 \\
InDuDoNet+~\cite{indudonet_plus} & 34.40 ± 4.02 & 0.9792 ± 0.0069 & 6.29 ± 2.96 & 0.0768 ± 0.0187 & 39.47 ± 3.61 & 0.9884 ± 0.0035 & 3.45 ± 1.60 & 0.0550 ± 0.0129 \\
OSCNet~\cite{oscnet} & 39.20 ± 3.00 & 0.9874 ± 0.0035 & 3.47 ± 1.37 & 0.0582 ± 0.0156 & 40.28 ± 4.05 & 0.9906 ± 0.0030 & 3.24 ± 1.86 & 0.0443 ± 0.0122 \\
OSCNet+~\cite{oscnet_plus} & 39.60 ± 3.06 & 0.9876 ± 0.0035 & 3.32 ± 1.36 & 0.0578 ± 0.0152 & 41.09 ± 3.77 & 0.9912 ± 0.0025 & 2.91 ± 1.63 & 0.0442 ± 0.0119 \\
MEPNet~\cite{mepnet} & 36.53 ± 3.07 & 0.9795 ± 0.0090 & 4.71 ± 1.70 & 0.0764 ± 0.0228 & 40.68 ± 3.45 & 0.9900 ± 0.0028 & 2.98 ± 1.34 & 0.0532 ± 0.0152 \\
DICDNet~\cite{dicdnet} & 38.21 ± 3.81 & 0.9850 ± 0.0045 & 4.06 ± 2.25 & 0.0692 ± 0.0169 & 40.01 ± 3.62 & 0.9899 ± 0.0026 & 3.26 ± 1.68 & 0.0519 ± 0.0127 \\
ACDNet~\cite{acdnet} & 38.24 ± 2.77 & 0.9869 ± 0.0036 & 3.84 ± 1.37 & 0.0606 ± 0.0158 & 40.06 ± 3.71 & 0.9910 ± 0.0022 & 3.26 ± 1.66 & 0.0461 ± 0.0116 \\
DuDoDp~\cite{dudop} & 43.25 ± 1.78 & 0.9878 ± 0.0034 & 2.09 ± 0.44 & 0.0689 ± 0.0110 & 44.34 ± 1.31 & 0.9900 ± 0.0023 & 1.82 ± 0.27 & 0.0577 ± 0.0086 \\
MARformer~\cite{marformer} & 43.73 ± 4.02 & 0.9889 ± 0.0060 & 2.15 ± 0.96 & 0.0740 ± 0.0281 & 47.15 ± 3.63 & 0.9936 ± 0.0024 & 1.42 ± 0.56 & 0.0506 ± 0.0136 \\
MARViT (ours) & 45.31 ± 1.64 & 0.9914 ± 0.0021 & 1.64 ± 0.32 & 0.0757 ± 0.0135 & 46.05 ± 1.90 & 0.9927 ± 0.0015 & 1.52 ± 0.32 & 0.0717 ± 0.0097 \\
MARPVT (ours) & 45.96 ± 1.17 & 0.9926 ± 0.0018 & 1.51 ± 0.21 & 0.0480 ± 0.0123 & 46.81 ± 2.45 & 0.9938 ± 0.0018 & 1.41 ± 0.38 & 0.0390 ± 0.0092 \\
MARMamba (ours) & \redbf{47.06} ± 1.58 & \redbf{0.9933} ± 0.0015 & \redbf{1.34} ± 0.24 & \redbf{0.0435} ± 0.0113 & \redbf{47.56} ± 3.17 & 0.9941 ± 0.0017 & \redbf{1.33} ± 0.46 & \redbf{0.0338} ± 0.0086 \\ \hline
\end{tabular}
}
\label{table:has_metal_large_medium}
\end{table*}

\begin{table*}[]
\caption{\rone{Quantitative comparison experiments. Evaluate the effectiveness of metal artifact restoration for \textbf{small} and \textbf{tiny} sizes using the PSNR($\uparrow$), SSIM($\uparrow$), RMSE($\downarrow$), LPIPS($\downarrow$) metrics \textbf{(metal-included region)}. Each grid cell represents the mean $\pm$ standard deviation. The optimal mean values highlighted in \redbf{bold red font}.}}
\centering
\resizebox{\textwidth}{!}{
\begin{tabular}{ccccc|cccc}
\hline
\multirow{2}{*}{Model} & \multicolumn{4}{c|}{Small} & \multicolumn{4}{c}{Tiny} \\
 & PSNR & SSIM & RMSE & LPIPS & PSNR & SSIM & RMSE & LPIPS \\ \hline
InDuDoNet~\cite{indudonet} & 43.81 ± 3.98 & 0.9937 ± 0.0022 & 2.25 ± 2.11 & 0.0410 ± 0.0101 & 46.57 ± 2.11 & 0.9949 ± 0.0009 & 1.44 ± 0.40 & 0.0383 ± 0.0091 \\
InDuDoNet+~\cite{indudonet_plus} & 45.95 ± 2.52 & 0.9935 ± 0.0015 & 1.56 ± 0.49 & 0.0446 ± 0.0105 & 46.98 ± 2.01 & 0.9942 ± 0.0011 & 1.37 ± 0.33 & 0.0416 ± 0.0098 \\
OSCNet~\cite{oscnet} & 46.02 ± 2.49 & 0.9944 ± 0.0013 & 1.55 ± 0.47 & 0.0364 ± 0.0097 & 47.10 ± 2.27 & 0.9950 ± 0.0009 & 1.36 ± 0.44 & 0.0343 ± 0.0096 \\
OSCNet+~\cite{oscnet_plus} & 46.26 ± 2.49 & 0.9945 ± 0.0013 & 1.51 ± 0.46 & 0.0364 ± 0.0099 & 47.30 ± 2.34 & 0.9950 ± 0.0009 & 1.33 ± 0.43 & 0.0342 ± 0.0097 \\
MEPNet~\cite{mepnet} & 45.80 ± 1.67 & 0.9934 ± 0.0013 & 1.55 ± 0.34 & 0.0461 ± 0.0131 & 47.98 ± 1.66 & 0.9943 ± 0.0011 & 1.21 ± 0.28 & 0.0433 ± 0.0126 \\
DICDNet~\cite{dicdnet} & 45.52 ± 2.51 & 0.9937 ± 0.0014 & 1.64 ± 0.50 & 0.0430 ± 0.0107 & 46.34 ± 2.24 & 0.9942 ± 0.0010 & 1.48 ± 0.44 & 0.0409 ± 0.0102 \\
ACDNet~\cite{acdnet} & 43.46 ± 3.00 & 0.9944 ± 0.0012 & 2.12 ± 0.74 & 0.0380 ± 0.0101 & 43.88 ± 3.31 & 0.9948 ± 0.0009 & 2.04 ± 0.78 & 0.0353 ± 0.0097 \\
DuDoDp~\cite{dudop} & 45.83 ± 1.09 & 0.9928 ± 0.0012 & 1.53 ± 0.20 & 0.0513 ± 0.0070 & 45.97 ± 1.03 & 0.9929 ± 0.0010 & 1.50 ± 0.18 & 0.0500 ± 0.0068 \\
MARformer~\cite{marformer} & 49.64 ± 1.33 & 0.9954 ± 0.0009 & 0.99 ± 0.17 & 0.0479 ± 0.0119 & 50.58 ± 1.15 & 0.9959 ± 0.0005 & 0.89 ± 0.15 & 0.0455 ± 0.0121 \\
MARViT (ours) & 48.64 ± 0.88 & 0.9946 ± 0.0008 & 1.10 ± 0.12 & 0.0715 ± 0.0106 & 49.16 ± 0.70 & 0.9948 ± 0.0004 & 1.04 ± 0.09 & 0.0689 ± 0.0088 \\
MARPVT (ours) & 49.84 ± 0.96 & 0.9958 ± 0.0008 & 0.96 ± 0.11 & 0.0361 ± 0.0081 & 50.55 ± 0.70 & 0.9961 ± 0.0005 & 0.88 ± 0.07 & 0.0342 ± 0.0076 \\
MARMamba (ours) & \redbf{50.76} ± 1.37 & \redbf{0.9958} ± 0.0009 & \redbf{0.87} ± 0.14 & \redbf{0.0320} ± 0.0072 & \redbf{51.99} ± 0.97 & \redbf{0.9963} ± 0.0005 & \redbf{0.75} ± 0.09 & \redbf{0.0280} ± 0.0061 \\ \hline
\end{tabular}
}
\label{table:has_metal_small_tiny}
\end{table*}

\rone{In addition, we conducted a comprehensive evaluation of the entire test set, including metallic regions, as detailed in Tables~\ref{table:has_metal_large_medium},~\ref{table:has_metal_small_tiny}. When analyzing metal artifact removal at large and medium scales with metallic areas included (Table~\ref{table:has_metal_large_medium}), all models displayed reduced performance compared to scenarios without metallic regions. Physically-informed prior-based methods, particularly InDuDoNet~\cite{indudonet} and InDuDoNet+~\cite{indudonet_plus}, experienced the most significant performance decline, with a PSNR drop of around 5 dB at the large scale, highlighting the added complexity posed by metallic regions. The diffusion-based DuDoDp~\cite{dudop} demonstrated greater resilience, with only a 0.22 dB reduction. Among Transformer-based models, MARformer~\cite{marformer} performed exceptionally well at the medium scale, while MARViT maintained stability but showed higher LPIPS values, and MARPVT achieved a balanced performance across numerical and perceptual metrics. Notably, MARMamba consistently outperformed all other models across scales, delivering excellent results with slightly higher but still low standard deviations. These findings further affirm MARMamba's robustness and extend its demonstrated superiority from non-metallic experiments to cases involving metallic regions.}

\rone{For artifact removal in regions with metal at small and tiny scales, all models showed improved performance compared to their results at large and medium scales. This suggests that artifacts from smaller metals are generally easier to restore. Our proposed model consistently delivered superior average restoration outcomes compared to other approaches. However, its standard deviation was still higher than those of MARViT and MARPVT, indicating a need for further refinement in stability. Overall, across artifacts from metals of varying sizes and their respective regions, the MARMamba model demonstrates state-of-the-art numerical accuracy while effectively balancing structural fidelity and perceptual quality.}

\begin{table}[]
\caption{\rone{Parameters vs. Inference Time}}
\centering
\tiny
\resizebox{\linewidth}{!}{
\begin{tabular}{ccc}
\hline
Model & Time/s & Param/M \\ \hline
InDuDoNet~\cite{indudonet} & 0.2537±0.0037 & 5.09 \\
InDuDoNet+~\cite{indudonet_plus} & 0.2521±0.0041 & 1.75 \\
OSCNet~\cite{oscnet} & 0.0181±0.0004 & 1.27 \\
OSCNet+~\cite{oscnet_plus} & 0.0243±0.0006 & 1.28 \\
MEPNet~\cite{mepnet} & 0.3585±0.0168 & 4.72 \\
DICDNet~\cite{dicdnet} & 0.0187±0.0007 & 1.28 \\
ACDNet~\cite{acdnet} & 0.0270±0.0011 & 1.60 \\
DuDoDp~\cite{dudop} & 2.9112±0.2489 & 82.36 \\
MARformer~\cite{marformer} & \redbf{0.0164}±0.0009 & 0.67 \\
MARViT (ours) & 0.0176±0.0006 & 0.57 \\
MARPVT (ours) & 0.0239±0.0017 & 0.62 \\
MARMamba (ours) & 0.0393±0.0008 & \redbf{0.59} \\ \hline
\end{tabular} 
}
\begin{flushleft}
    Note: \redbf{Bold red text} indicates the optimal value.
  \end{flushleft}
  \vspace{-0.3em}
\label{table:time_param}
\end{table}

\rone{In addition to the quantitative evaluation of CT metal artifact removal, we assessed inference efficiency and parameter scale (Table \ref{table:time_param}). The results highlight distinct differences among the models. Physics-based methods, such as InDuDoNet~\cite{indudonet} and MEPNet~\cite{mepnet}, exhibit larger parameter sizes and longer runtime requirements. In contrast, OSCNet~\cite{oscnet} and DICDNet~\cite{dicdnet} demonstrate quicker execution times. However, the diffusion-based DuDoDp~\cite{dudop}, constrained by its design, requires significantly more parameters and inference time compared to all other models. Transformer-based architectures strike a more balanced performance between efficiency and effectiveness. Among them, MARformer~\cite{marformer} achieves the fastest inference, reducing runtime by 0.0229 seconds compared to MARMamba. While marginally slower, MARMamba uses only 0.59 million parameters, the smallest parameter count across all models, highlighting its exceptional efficiency alongside its strong artifact removal capability.}

\begin{figure*}[h]
\centering
\includegraphics[width=\textwidth]{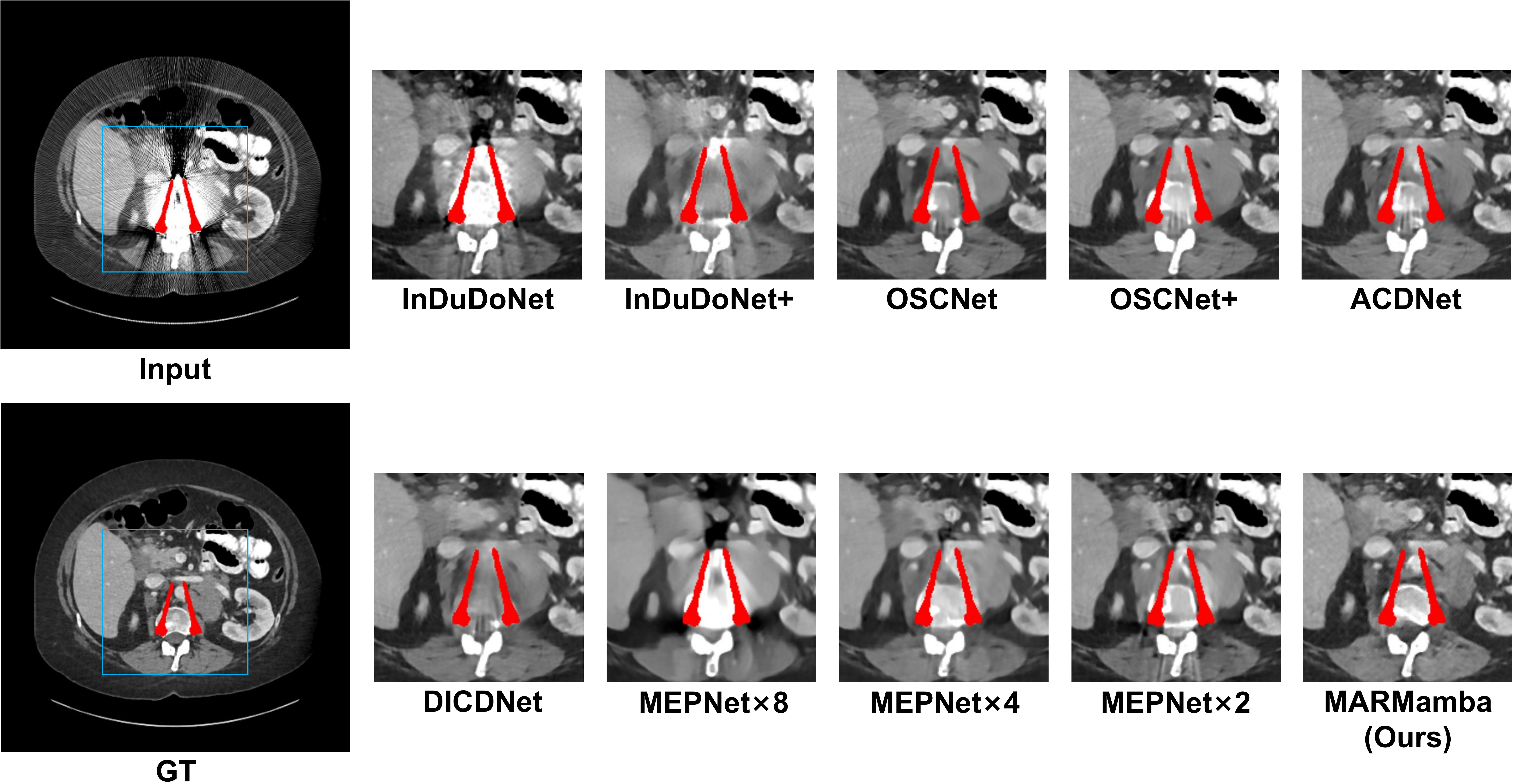}
\caption{\revise{Large-scale visual comparison of metallic implants. The input image is selected from the SynDeepLesion~\cite{indudonet} dataset. The red regions in the images indicate the locations of metallic components. To better visualize fine details, only the content within the blue bounding boxes is displayed for each model. A Hounsfield Unit (HU) window of [-175 HU, 275 HU] is used for visualization.}} 
\label{fig:large}
\end{figure*}

\revise{Fig. \ref{fig:large} provides a visual comparison of different CT MAR models in addressing the artifacts caused by large metallic implants. These images are sourced from the SynDeepLesion~\cite{indudonet} test set. Artifacts from large-scale metal implants severely compromise the quality of CT images. Upon examining the enlarged sections of each model, it becomes evident that both InDuDoNet~\cite{indudonet} and InDuDoNet+~\cite{indudonet_plus} struggle to effectively suppress the intense artifacts caused by large metallic objects, as distinct streaks remain apparent. In contrast, models such as OSCNet~\cite{oscnet}, ACDNet~\cite{acdnet}, and DICDNet~\cite{dicdnet} offer more effective artifact suppression. However, they often compromise structural details and fine textures in the surrounding tissue. Additionally, OSCNet+~\cite{oscnet_plus} and MEPNet~\cite{mepnet} introduce significant blurring in both soft tissue and metal-covered regions. Our proposed model, MARMamba, excels in several areas by effectively reducing severe metal-induced artifacts while maintaining anatomical details and generating more natural image textures. The restored images closely resemble artifact-free references. These results underscore the significant advantage of MARMamba in managing challenging artifact removal scenarios associated with large metallic implants.}

\begin{figure*}[h]
\centering
\includegraphics[width=\textwidth]{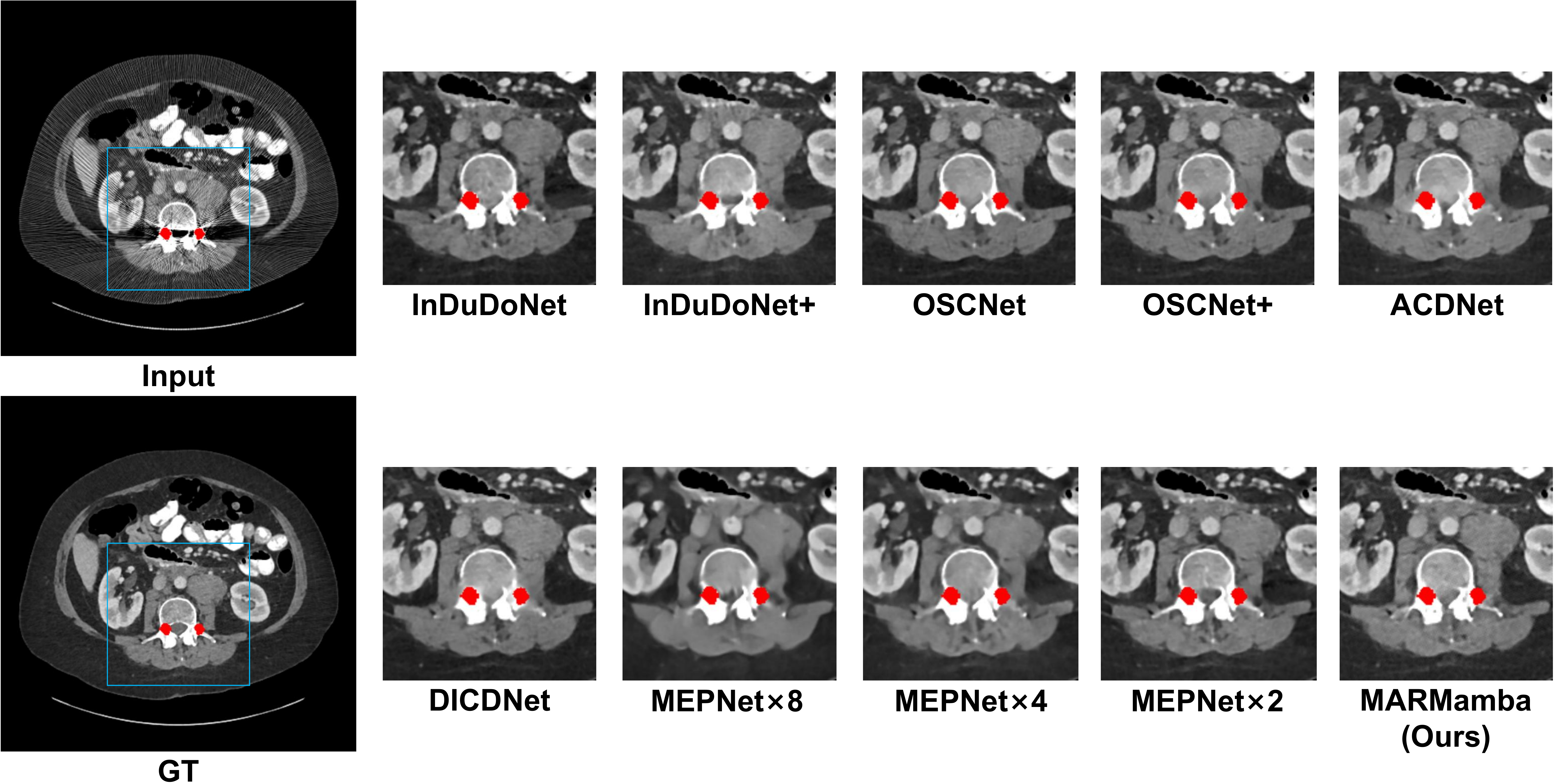}
\caption{\revise{Medium metallic implants. The input image is selected from the SynDeepLesion~\cite{indudonet} dataset. The HU window is set to the range of [-175 HU, 275 HU].}}
\label{fig:medium}
\end{figure*}

\revise{Medium-sized metallic implants cause more localized artifacts and less severe degradation to soft tissue structures compared to larger implants. Fig. \ref{fig:medium} illustrates how different CT metal artifact reduction models tackle artifacts from medium-sized metallic implants. Although InDuDoNet~\cite{indudonet} and InDuDoNet+~\cite{indudonet_plus} show progress in reducing artifacts near the central metallic area, they are less effective along the implant boundaries. Models like OSCNet~\cite{oscnet}, ACDNet~\cite{acdnet}, and DICDNet~\cite{dicdnet} are relatively successful in suppressing streak artifacts. However, they often cause noticeable smoothing, which diminishes edge sharpness and soft tissue texture, leading to some structural blurring. Meanwhile, OSCNet+~\cite{oscnet_plus} and the MEPNet~\cite{mepnet} series perform less effectively in high-contrast edge areas, resulting in over-smoothing and decreased visual clarity. In contrast, our proposed model, MARMamba, effectively removes artifacts while better preserving soft tissue structures near metallic boundaries. Nonetheless, its capacity to restore high-frequency edge details in high-contrast regions is limited, posing a constraint of the current model.}

\begin{figure*}[h]
\centering
\includegraphics[width=\textwidth]{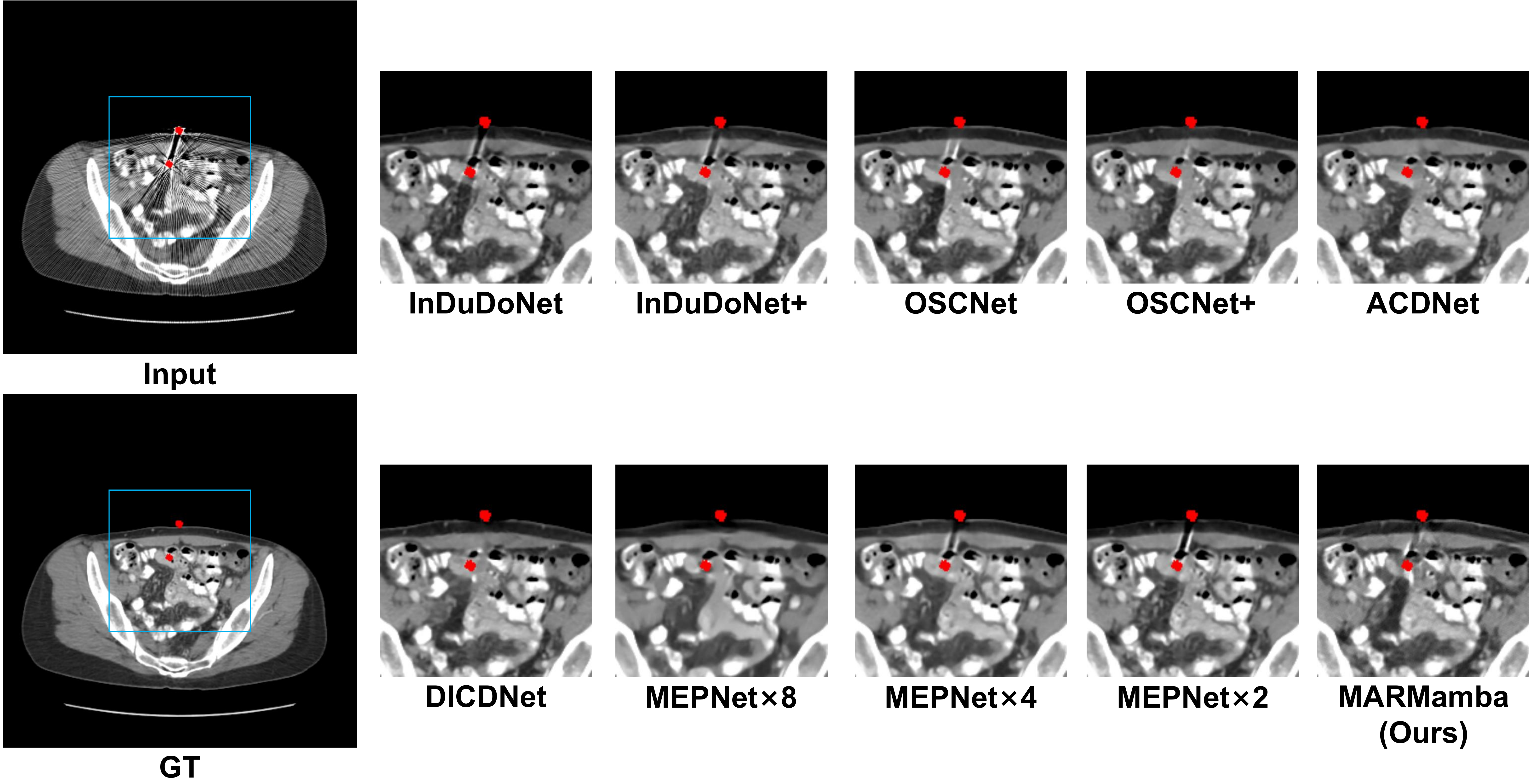}
\caption{\revise{Small metallic implants. The input image is selected from the SynDeepLesion~\cite{indudonet} dataset. The HU window is set to the range of [-175 HU, 275 HU].}}
\label{fig:small}
\end{figure*}

\revise{Small metallic implants create artifacts that are spatially confined but primarily disrupt fine anatomical details. Fig. \ref{fig:small} showcases the restoration performance of various models in addressing these artifacts. InDuDoNet~\cite{indudonet}, InDuDoNet+~\cite{indudonet_plus}, and MEPNet~\cite{mepnet} effectively remove artifacts at a distance but perform suboptimally in low-contrast regions, especially near the implant center, where some texture distortions remain. OSCNet~\cite{oscnet}, ACDNet~\cite{acdnet}, and DICDNet~\cite{dicdnet} can effectively reduce localized metal artifacts but tend to over-smooth small-scale structures, leading to blurred tissue boundaries. In contrast, our proposed model, MARMamba, excels in recovering structural details in settings with small metal artifacts. It effectively removes localized artifacts around the implant and suppresses distant distortions while preserving the sharpness of soft tissue separations and bone edges. The overall texture of the restored image more closely resembles that of an artifact-free reference. However, similar to InDuDoNet~\cite{indudonet} and InDuDoNet+~\cite{indudonet_plus}, MARMamba shows limited effectiveness in low-contrast regions, where some residual streaks remain visible near the center of the metallic implant in the image.}

\begin{figure*}[h]
\centering
\includegraphics[width=\textwidth]{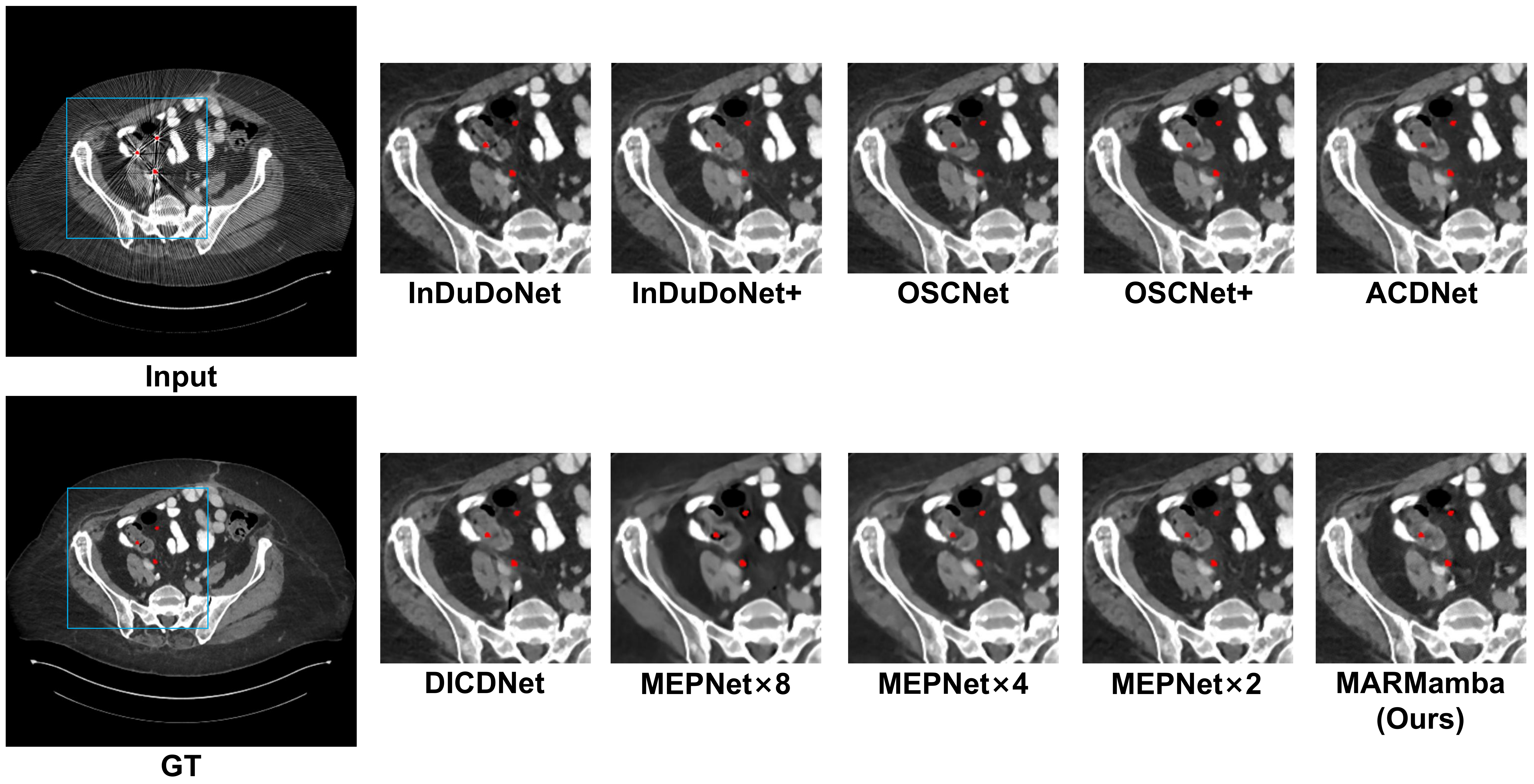}
\caption{\revise{Tiny metallic implants. The input image is selected from the SynDeepLesion~\cite{indudonet} dataset. The HU window is set to the range of [-175 HU, 275 HU].}}
\label{fig:tiny}
\end{figure*}

\revise{Artifacts from tiny metallic implants cause minimal overall image degradation but significantly affect fine-grained clinical structures, such as small vessels, trabecular bone, and soft tissue interfaces. Fig. \ref{fig:tiny} offers a comparative evaluation of various models in restoring CT images impacted by tiny metallic implants. Overall, all models perform well in reducing the streak artifacts these implants cause and succeed in maintaining the anatomical integrity of the surrounding tissues. However, it is important to highlight that minor deviations are still noticeable in areas of extremely low contrast in the output of the proposed model, MARMamba. Addressing these deviations is identified as a key area for future improvement.}

\begin{table}[]
\caption{\rone{Real‑World Scene Restoration Effectiveness Rating (1–5 Scale)}}
\centering
\normalsize
\begin{tabular*}{0.8\linewidth}{@{\extracolsep{\fill}}cc}
\hline
Model & Score \\ \hline
InDuDoNet~\cite{indudonet} & 3.489±1.223 \\
InDuDoNet+~\cite{indudonet_plus} & 3.586±1.194 \\
OSCNet~\cite{oscnet} & 3.409±1.252 \\
OSCNet+~\cite{oscnet_plus} & \redbf{3.863}±1.120 \\
ACDNet~\cite{acdnet} & 3.806±1.217 \\
MEPNet~\cite{mepnet} & 3.294±1.310 \\
DICDNet~\cite{dicdnet} & 3.657±1.279 \\
MARformer~\cite{marformer} & 3.680±1.296 \\
MARViT (ours) & 3.606±1.352 \\
MARPVT (ours) & 3.777±1.299 \\
MARMamba (ours) & 3.748±1.271 \\ \hline
\end{tabular*}
\begin{flushleft}
    \rtwo{Note: A total of 350 randomly selected samples were assessed, with the model identities concealed during scoring to prevent bias and enable an objective comparison.}
  \end{flushleft}
  \vspace{-0.3em}
\label{table:real}
\end{table}

\begin{figure*}[]
\centering
\includegraphics[width=\textwidth]{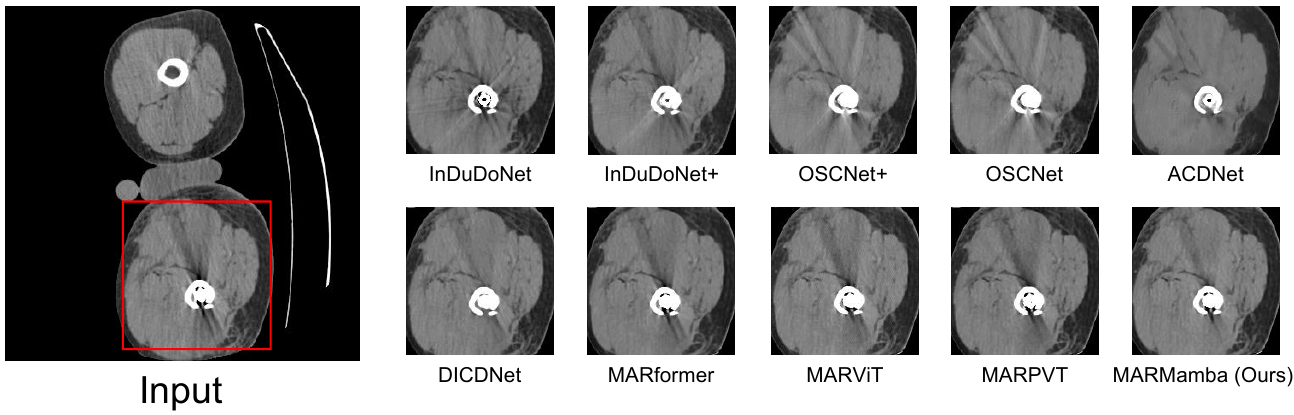}
\caption{Visual comparison in real-world scenarios. Excluding the input image, the visual images of the other models display the inner regions marked by the red rectangles. The display window is [-175HU, 275HU].} 
\label{fig:real}
\end{figure*}

\rone{Real-world CT images exhibit considerable variability due to differences in metal types, implantation methods, and anatomical contexts. These factors pose greater challenges for models in terms of generalization and detail reconstruction compared to synthetic scenarios.} To assess real-world performance, we used the CLINIC-metal~\cite{CTPelvic1K} dataset for comparison. \rtwo{As the dataset lacked rigorously aligned ground-truth references, it was not possible to calculate quantitative metrics. Therefore, a subjective evaluation was performed on 350 randomly selected samples. All authors participated in this evaluation, utilizing a five-point scoring system. In this system, a score of 1 indicates severe metal artifacts with significant structural distortion, while a score of 5 represents excellent artifact removal with well-preserved anatomical structures. To ensure fairness, the model identities were anonymized during the scoring process, and all samples were displayed using a consistent HU window range of [-175HU, 275HU]. The mean scores and standard deviations obtained from this evaluation are summarized in Table \ref{table:real}.}

\rone{The evaluation results indicate that OSCNet+~\cite{oscnet_plus} and ACDNet~\cite{acdnet} achieved the highest overall performance. This finding suggests that models explicitly designed to incorporate metal-artifact feature modeling demonstrate stronger generalization capabilities compared to both our method and transformer-based approaches, highlighting a limitation of our framework. Furthermore, all models exhibited standard deviations greater than one. This outcome can be attributed to the complexity and variability of large-scale metal artifacts present in real clinical settings, which synthetic data fails to fully replicate. Consequently, most models experience significant performance degradation when faced with these challenging scenarios.}

\rone{Fig. \ref{fig:real} showcases examples of real-world restoration results generated by each model. InDuDoNet~\cite{indudonet}, InDuDoNet+~\cite{indudonet_plus}, and OSCNet~\cite{oscnet} tend to leave noticeable residual artifacts and introduce smoothing effects on nearby anatomical structures. While OSCNet+~\cite{oscnet_plus} achieved the highest overall score in the evaluation, its performance on this specific type of artifact was not optimal. ACDNet~\cite{acdnet} is more effective at suppressing artifacts, but it also causes the image to appear overly smooth, which can obscure some of the original tissue details. Our model, along with other models based on Transformer architectures such as MARformer~\cite{marformer}, MARViT, and MARPVT, are better at preserving overall structural consistency. However, they are less effective at maintaining edge fidelity and suppressing artifacts in low-contrast regions.}

\subsection{Computation and Memory Analysis.}

\begin{figure}[]
\centering
\includegraphics[width=\linewidth]{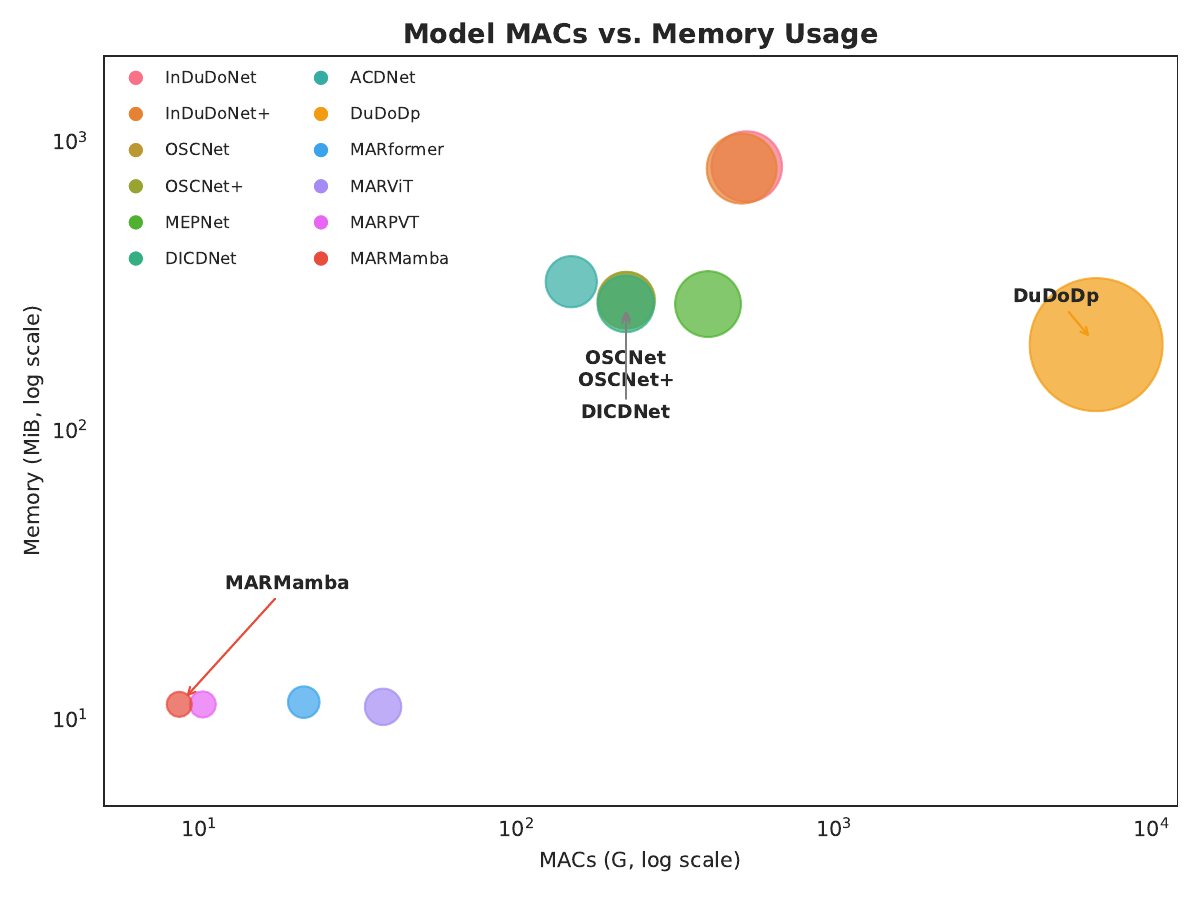}
\caption{\rtwo{Comparison of computational complexity and memory overhead. Bubble size is proportional to $\sqrt{\text{MACs}}$. Disable gradient during evaluation.}}
\label{fig:mac_memory}
\end{figure}

\rone{To evaluate the computational requirements of the models, we examined their memory usage and computational load under specific inference conditions. During the inference process, tensors with dimensions \(1 \times 1 \times 416 \times 416\) were used. For some models, such as InDuDoNet~\cite{indudonet}, additional input tensor sizes were consistent with those specified in the inference code. Computational load was assessed using Multiply-Accumulate (MAC) operations, measured in giga (G), while memory consumption was recorded in MiB. The results are illustrated in a bubble chart, as seen in Fig.~\ref{fig:mac_memory}}.

\rone{Physics-based models such as InDuDoNet~\cite{indudonet}, OSCNet~\cite{oscnet}, and MEPNet~\cite{mepnet} are positioned in the upper-right region of the diagram, indicating significantly higher computational complexity and memory demands. The larger bubble sizes associated with these models further emphasize their substantial resource requirements. In contrast, MARMamba and other Transformer-based architectures are grouped in the lower-left quadrant, marked by considerably lower computational and memory demands, with smaller bubbles reflecting their lightweight and efficient design. Notably, DuDoDp~\cite{dudop} stands out for its diffusion-based architecture, which involves multiple iterative steps, resulting in the highest computational burden—its primary limitation.}

\begin{figure}[htbp]
\centering
\includegraphics[width=\columnwidth]{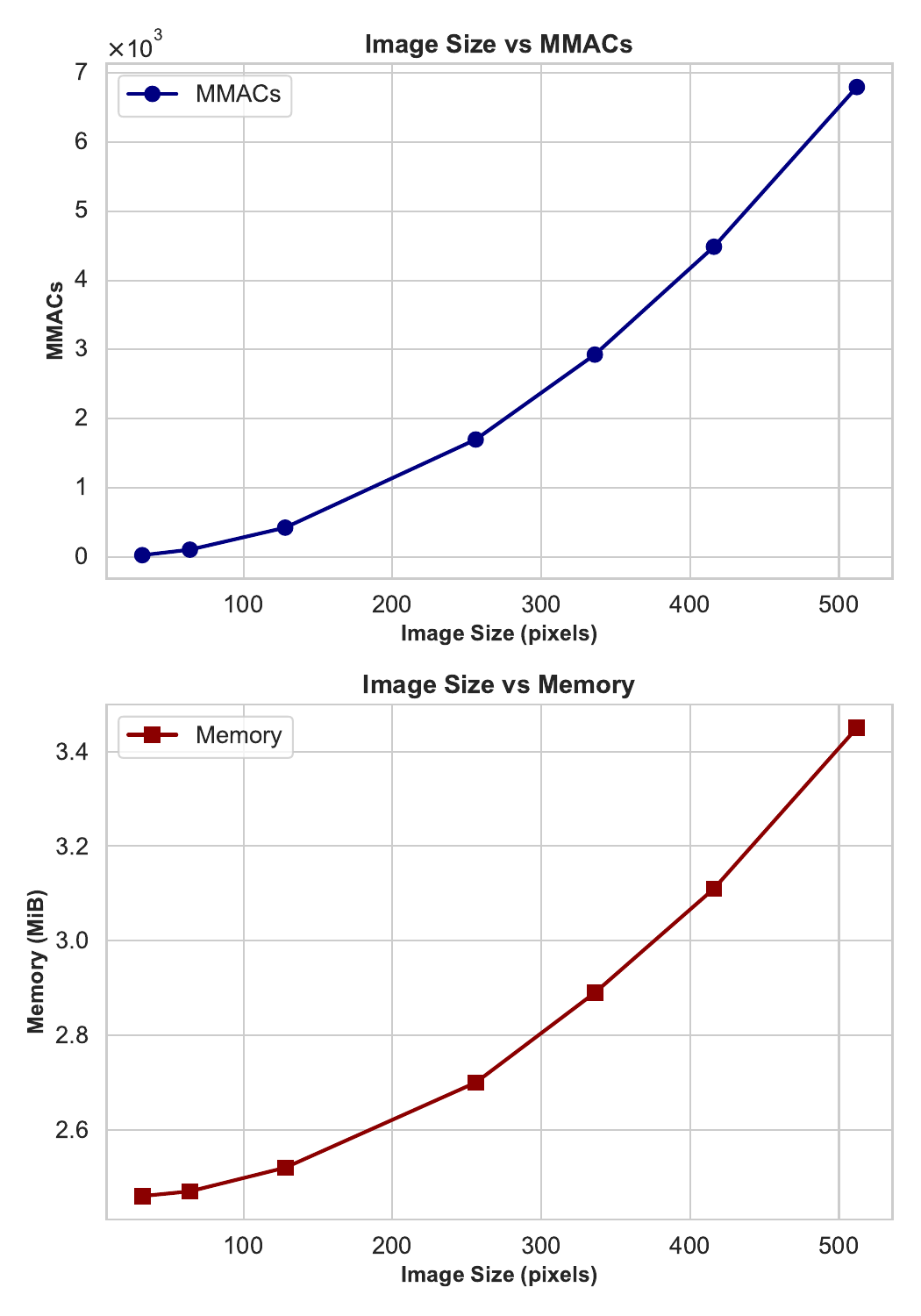}
\caption{\rone{Image size vs. resource consumption. Disable gradient during testing.}}
\label{fig:img_resource}
\end{figure}

\rone{Fig. \ref{fig:img_resource} highlights the relationship between input image size and resource usage, specifically model complexity and memory consumption. As the size of the input image increases, computational complexity exhibits an almost exponential growth, while memory usage rises at a much slower rate. For instance, when image dimensions expand from a few dozen pixels to over 400 $\times$ 400, MMAC operations surge from tens to several thousand, illustrating the extreme sensitivity of computational complexity to resolution. In contrast, GPU memory consumption increases by only about 1 MiB. This contrast emphasizes that computational demands, rather than memory usage, represent the primary performance bottleneck. Therefore, future optimization efforts should prioritize enhancing computational efficiency.
}

\modify{In conclusion, the MARMamba model successfully balances computational efficiency with memory usage, matching the performance of optimized, lightweight Transformer-based architectures. Additionally, it excels in fine-grained restoration tasks, underscoring the significant advantages of the MARMamba model.}

\subsection{Ablation Study}

\begin{table}[htbp]
\caption{Ablation experiments within the FMB. \checkmark indicating enablement of each branch in the orientation scanning stage. \redbf{Red bold} font indicates the best value.}
\label{table:flip}
\centering
\resizebox{\linewidth}{!}{%
\begin{tabular}{ccclcc}
\hline
\multicolumn{1}{l}{Normal} & \multicolumn{1}{l}{Horizontal} & \multicolumn{1}{l}{Vertical} & PSNR/SSIM/RMSE & \multicolumn{1}{l}{Param/M} & \multicolumn{1}{l}{Time/ms} \\ \hline
\checkmark &  &  & 48.2703/0.9938/1.2469 & {\color[HTML]{FF0000} \textbf{0.47}} & {\color[HTML]{FF0000} \textbf{32.4}} \\
\checkmark & \checkmark &  & 48.6277/0.9941/1.1908 & 0.53 & 34.1 \\
\checkmark & \checkmark & \checkmark & {\color[HTML]{FF0000} \textbf{49.2559/0.9948/1.0784}} & 0.59 & 36.2 \\ \hline
\end{tabular}
}
\end{table}

\begin{figure*}[htbp]
\centering
\includegraphics[width=\textwidth]{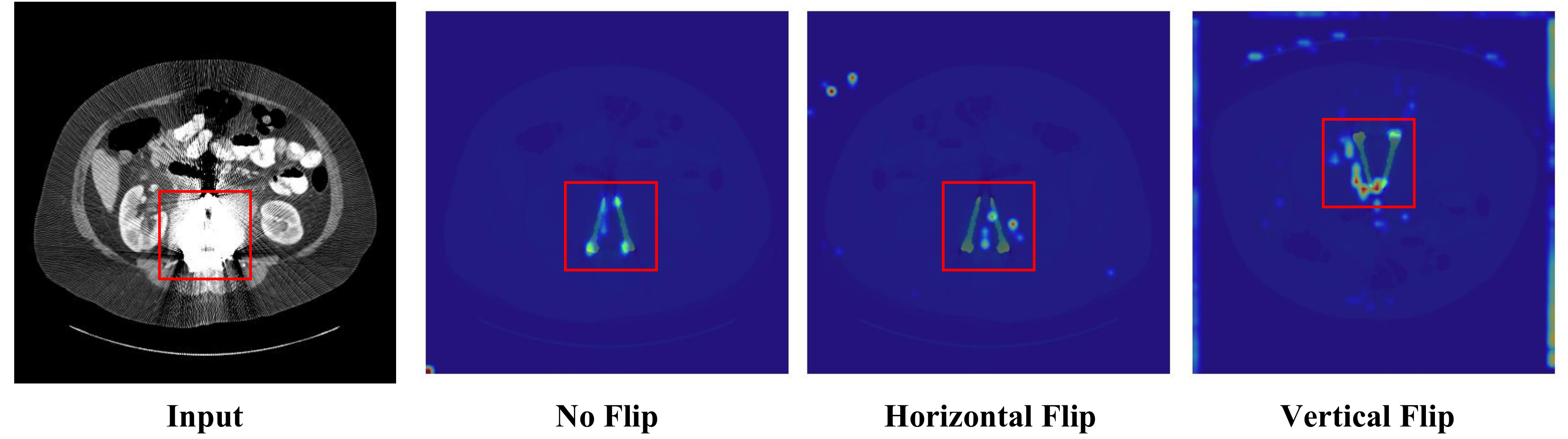}
\caption{\rone{Feature visualization of the FMB. From left to right, the panels illustrate the input image, the Mamba output corresponding to the original (unflipped) input, the output of the Mamba block under horizontal flipping, and the output under vertical flipping. The regions enclosed by red rectangles indicate the locations of metallic implants, which serve as the primary sources of artifacts. The display window of input image is [-175HU, 275HU].} }
\label{fig:FMB_visual}
\end{figure*}

\begin{table}
\caption{Ablation experiments within the AMB. \checkmark indicating enablement of each pooling layer. \redbf{Red bold} text highlights the optimal value.}
\centering
\resizebox{\linewidth}{!}{
\begin{tabular}{cclcc}
\hline
\multicolumn{1}{l}{Average} & \multicolumn{1}{l}{Maximum} & PSNR/SSIM/RMSE & \multicolumn{1}{l}{Time/ms} & \multicolumn{1}{l}{} \\ \hline
\checkmark &  & 48.5776/0.9940/1.2013 & 36.6 & \\
 & \checkmark & 48.4413/0.9940/1.2337 & {\color[HTML]{FF0000} \textbf{35.3}} & \\
\checkmark & \checkmark & {\color[HTML]{FF0000} \textbf{49.2559/0.9948/1.0784}} & 36.2 & \\ \hline
\end{tabular}
}
\label{table:AMFN}
\end{table}

To validate the effectiveness of the FMB and AMB modules, we performed ablation experiments using a full-image evaluation to ensure a comprehensive performance assessment on their internal structures, summarized in Tables~\ref{table:flip} and \ref{table:AMFN}. For the FMB module, we evaluated different branches within the orientation scanning phase individually, as detailed in Table~\ref{table:flip}. Enabled three branches significantly improved the model’s restoration performance. Compared to using only the branch without flipping the feature map, activating all branches increased the PSNR by approximately 2\% and reduced the RMSE by around 14\%. However, this improvement came with a slight increase in resource requirements, adding 0.12 M parameters and an additional 3.8 ms to the inference time. This modest trade-off in computational cost is justified by the notable gains in effectiveness and performance. 

\rone{We utilized Gradient-weighted Class Activation Mapping (Grad-CAM) to analyze the outputs of the three Mamba blocks within the FMB module of MS-Mamba, following the network's downsampling stage. The resulting visualizations are shown in Fig.~\ref{fig:FMB_visual}. In the original input, the model's attention is primarily focused on the region containing the metallic artifact. However, in the Mamba blocks processing the vertically and horizontally flipped images, the model adaptively shifts its focus to the corresponding positions of the metal. These findings indicate that the MS-Mamba module does not rely on fixed spatial coordinates for its responses. Instead, it shows directional sensitivity and structural awareness of the image content. The module dynamically adjusts its activation patterns based on changes in input orientation, demonstrating that the FMB design enables MS-Mamba to effectively capture information across different spatial orientations.}

\begin{figure}[htbp]
\centering
\includegraphics[width=\linewidth]{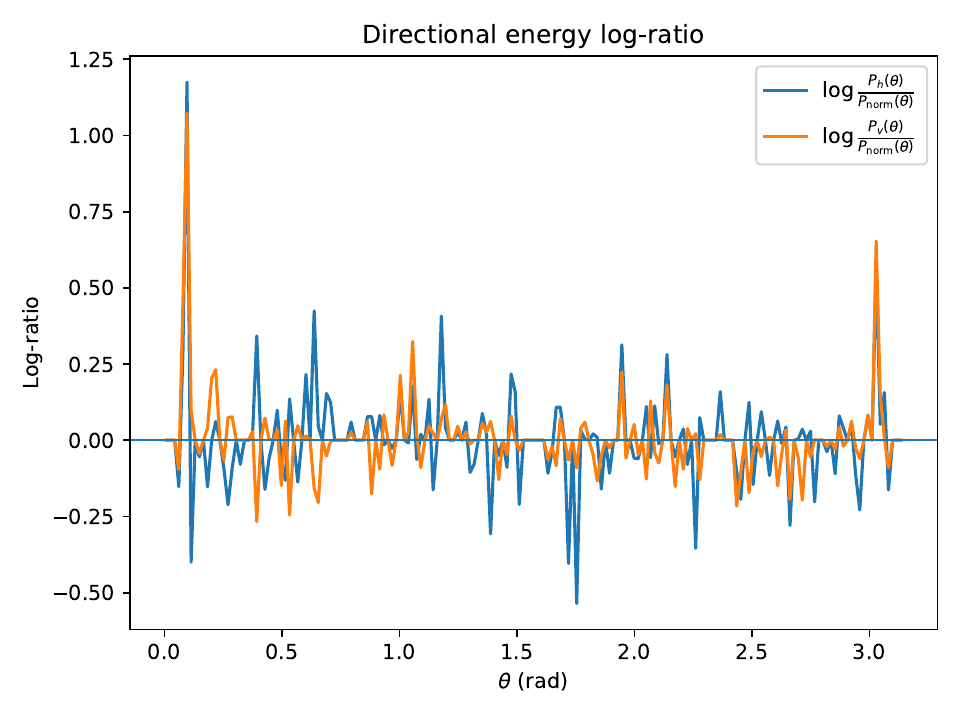}
\caption{\rtwo{Directional energy log-ratio of MS-Mamba branches. Positive values indicate orientations where the flipped branches selectively enhance structural responses relative to the non-flipped branch, while negative values indicate suppression.}}
\label{fig:directional_energy_logratio}
\end{figure}

\rtwo{To assess the model's ability to capture orientation-specific features, we conducted a directional energy analysis in the frequency domain on the features extracted from different branches of the FMB. This analysis involved applying a two-dimensional Fourier transform to the feature maps from each branch and calculating the directional energy distribution, \( P(\theta) \), within a mid-frequency annular band. This band predominantly represents the streak-like structures caused by metal artifacts. Subsequently, we computed the logarithmic ratio of directional energy between each flipped branch and the corresponding non-flipped branch using the formula \( \log\!\left(P_{\mathrm{branch}}(\theta)/P_{\mathrm{norm}}(\theta)\right) \). The directional energy log-ratio curves derived from this analysis are shown in Fig.~\ref{fig:directional_energy_logratio}.}

\rtwo{In this analysis, a log-ratio value of zero indicates that the model's response at a given orientation is identical to that of the non-flipped branch. Positive values signify that the flipped branch enhances structures at a specific orientation, while negative values indicate suppression. The results reveal that for most orientations, the log-ratio remains near zero, suggesting the model maintains a stable and largely isotropic response overall. However, at certain distinct orientations, the horizontally flipped and vertically flipped branches exhibit pronounced positive peaks, with each branch showing peaks at different locations. This behavior demonstrates that the flipped branches selectively enhance structural responses at complementary orientations, rather than uniformly amplifying all directions. These findings validate that the proposed MS-Mamba effectively models orientation-specific information in a controlled and complementary manner at the feature level.
}

Table~\ref{table:AMFN} indicates the impact of activating the average pooling or maximum pooling layers. Both pooling layers avoids unnecessary computational overhead while delivering performance benefits. Specifically, the combination improves PSNR by 0.68 dB compared to using only the average pooling layer and by 0.82 dB compared to using only the maximum pooling layer. Together, these findings confirm that the FMB and AMB modules achieve a well-balanced trade-off between performance and efficiency.

\begin{table}[]
\centering
\caption{\rone{Loss Function Ablation Experiment.}}
\setlength{\tabcolsep}{20pt}
\begin{tabular}{cc}
\hline
Loss & PSNR$\uparrow$/SSIM$\uparrow$/RMSE$\downarrow$ \\ \hline
Pseudo-Huber & 48.4497/0.9944/1.1849 \\
LPIPS & 46.1632/0.9931/1.5145 \\
Pseudo-Huber+LPIPS & {\color[HTML]{FE0000} \textbf{49.2559/0.9948/1.0784}} \\ \hline
\end{tabular}
\label{table:loss}
\end{table}

\rone{The results of the ablation experiments on loss functions are presented in Table~\ref{table:loss}. Using the Pseudo-Huber loss alone delivers strong performance, underscoring its effectiveness in minimizing pixel-level differences. In contrast, relying solely on the LPIPS loss produces poorer results across all three metrics compared to Pseudo-Huber, suggesting that while LPIPS prioritizes perceptual similarity, it falls short in numerical accuracy. Combining both loss functions achieves the best overall performance. Pseudo-Huber ensures precise and stable numerical consistency, while LPIPS incorporates perceptual constraints that enhance visual quality, making the results more suitable for clinical interpretation.}

\section{Discussion}

\begin{figure}[htbp]
  \centering
  \includegraphics[scale=0.3]{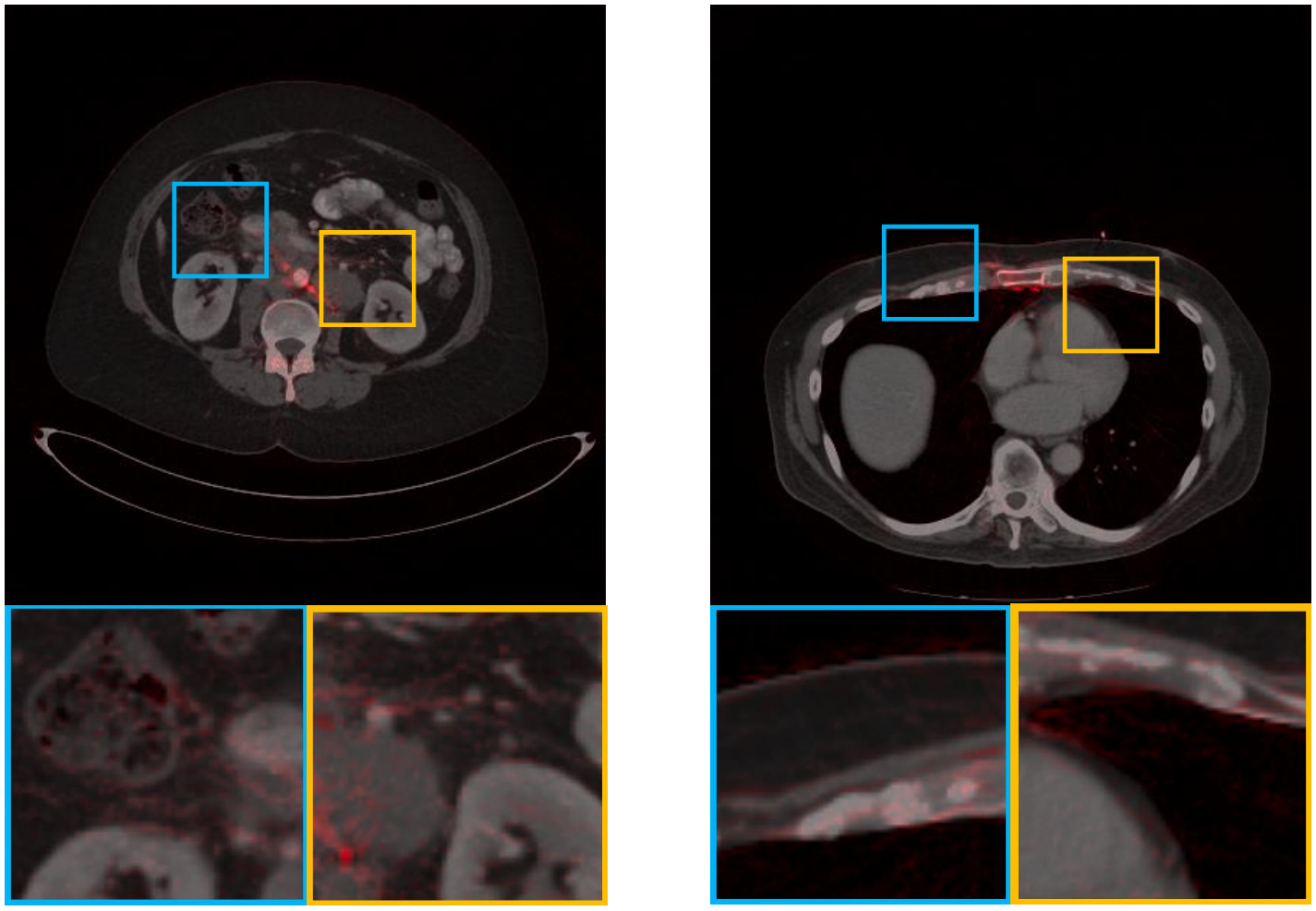}
  \caption{\rone{Error maps between restored images and ground truths. Samples selected from the SynDeepLesion test set~\cite{indudonet}. Magnified details are provided below each image for observation. Red indicates the absolute error between the repaired image and the ground truth. To highlight the errors, the error values have been amplified by a factor of two. The metal-containing area was preserved in the restored image. HU window: [-175 HU, +275 HU].}}
  \label{fig:error_map}
\end{figure}

The experimental analysis indicates that the proposed MARMamba model effectively removes metal artifacts across various implant sizes while maintaining high fidelity in preserving detailed anatomical and organ-level textures. Additionally, MARMamba strikes a favorable balance between restoration performance and computational efficiency, with its parameter count, memory usage, and computational costs similar to those of other lightweight models. \rone{Achieving this balance allows our model to be effectively deployed on resource-constrained platforms, such as embedded devices used in image acquisition systems. In such applications, clinicians receive both the enhanced and original images for side-by-side comparison, aiding in more accurate lesion evaluation and clinical diagnosis. However, despite its advantages, the proposed model has several limitations, which are discussed below:}
\begin{itemize}
    \item \revise{\textbf{Restoration in low-contrast regions.} Suboptimal restoration in low-contrast regions remains a significant challenge. As demonstrated in Figs. \ref{fig:medium}, \ref{fig:small}, and \ref{fig:tiny}, our model struggles to fully eliminate streak artifacts in areas of low contrast. Additionally, the structural details recovered by the model show noticeable deviations from the artifact-free references. This limitation is largely due to the model's inadequate ability to effectively integrate multi-contrast information. \rone{To illustrate the limitations of MARMamba more clearly, Fig.~\ref{fig:error_map} displays the error map between the restored image and the ground truth. Upon magnifying specific sections, it becomes evident that apart from the metal-containing region, errors are concentrated around low-contrast tissues where grayscale differences are negligible and artifact intensities closely mimic real tissue. In these areas, feature extraction faces challenges, resulting in decreased separability and noticeable prediction bias. This issue mainly stems from the loss function's inadequacy in effectively constraining low-contrast regions. The Pseudo-Huber loss, which measures pixel-wise intensity differences, struggles in areas where grayscale variations are subtle, making it less responsive. Similarly, the LPIPS loss relies on features extracted by deep learning models, but texture and edge features are naturally weak in low-contrast zones. Consequently, the differences detected by LPIPS are not sufficiently distinct, further limiting its effectiveness.}}
    \item \rone{\textbf{Potential biases in training dataset.} The training dataset, consisting primarily of synthetic images, fails to capture the full diversity of real implants. This limitation introduces biases during the training process, including imbalanced sample distributions and artificial artifacts that do not fully represent the complexity of real-world metal artifacts. As a result, the model's generalizability and performance in clinical applications are restricted.}
\end{itemize}

\rone{To address the limitations of our approach, we propose several avenues for improvement in future work, as outlined below:} 
\begin{itemize}

    \item \rone{\textbf{Low-contrast enhancement.} We plan to develop a contrast-aware feature interaction module that leverages multi-contrast information. This enhancement aims to improve the restoration of anatomical details, particularly in low-contrast regions.}

    \item \rone{\textbf{Robustness improvement.} To enhance the generalizability of our model, we plan to expand the diversity of the training dataset by collecting real implant images that vary in type, size, and material. In addition, we will adopt a hybrid training strategy that integrates synthetic data with clinically acquired images. To further ensure a comprehensive assessment of model performance, expert radiologist scoring will be incorporated as part of the evaluation protocol. This approach will leverage the scalability of synthetic data while ensuring the clinical relevance of real-world data.}

    \item \rone{\textbf{3D volumes support.} To address volumetric artifacts, we aim to design a hybrid architecture that incorporates 3D convolutional and pooling layers. This architecture will refine the loss function for 3D volumes and use cone-beam projection domain guidance to maintain consistency along the z-axis.}

    \item \rtwo{\textbf{CUDA optimization.} In terms of computational efficiency, our model's current inference time is slightly longer compared to some Transformer-based methods. This is mainly due to the additional feature transformations and data-access operations required by the multi-branch architecture. To address this issue, future work will concentrate on optimizing implementation techniques. These include fusing consecutive operations to minimize redundant memory access, designing more compact branch implementations to eliminate unnecessary computations while retaining directional modeling capabilities, using mixed-precision inference to leverage modern GPU hardware more effectively, and employing efficient operator implementations along with hardware-aware optimizations to reduce end-to-end latency without changing the network architecture.}

    \item\rone{\textbf{Physics‑informed modeling and loss design.} Future work will incorporate additional physics‑based constraints into both the feature representation and loss formulation. On the feature side, we will integrate physical priors such as projection characteristics in the frequency domain and attenuation properties reflected by regional intensities, embedding physical‑level constraints directly into the image‑domain modeling process. On the loss side, we will extend the current formulation by introducing frequency‑aware penalties and components tailored to CT‑specific phenomena, including non‑stationary noise distributions and beam‑hardening effects near metal implants. Together, these enhancements are expected to improve the model’s robustness, interpretability, and alignment with underlying CT imaging physics.}
    
\end{itemize}

\rone{These planned improvements are expected to significantly enhance the practical applicability and reliability of our MARMamba model, positioning it for broader deployment in real-world medical settings.}

\section{Conclusion}

\modify{In this paper, we introduce MARMamba, a streamlined model derived from the Mamba architecture, engineered to efficiently remove multi-scale metal artifacts from CT images. It maintains a balance between restoration quality, computational cost, and resource usage. The model processes CT images with metal artifacts directly, without relying on additional inputs like sinogram domain data or masks.} MARMamba utilizes a UNet-like architecture with MS-Mamba as the core module. MS-Mamba captures multi-orientation information via FMB and fuses salient and average features using AMFN. These techniques enable MARMamba to effectively eliminate multi-scale metal artifacts while maintaining low computational overhead. Experimental results show that the MARMamba model surpasses existing methods in quality of metal artifact removal. Additionally, it matches the resource efficiency of lightweight Transformer-based architectures regarding computational load, memory usage, and parameter count. These findings indicate that MARMamba effectively balances restoration performance with resource efficiency, underscoring its practical applicability. \rone{Our model demonstrates potential for broad application in medical imaging, particularly within dentistry, orthopedics, and cardiovascular imaging. In orthopedics, the model could minimize artifacts caused by hip or knee prostheses, facilitating more accurate assessments of bone healing and the detection of infections. Similarly, in dentistry, it offers the capacity to reduce implant-related artifacts in CT scans, enabling clearer evaluation of conditions such as periodontitis, cysts, and tumors.}

\section* {ETHICAL STATEMENTS}

This work did not involve human subjects or animals in its research.

\section*{ACKNOWLEDGMENT}

All authors declare that they have no known conflicts of interest in terms of competing financial interests or personal relationships that could have an influence or are relevant to the work reported in this paper. 

{
\bibliographystyle{IEEEtran}
\bibliography{IEEEabrv,references}
}

\end{document}